%% file: 01main.tex
\pdfoutput=1

\documentclass[lettersize,journal]{IEEEtran}
\usepackage{amsmath,amsfonts}
\usepackage{algorithmic}
\usepackage{algorithm}
\usepackage{array}
\usepackage[caption=false,font=normalsize,labelfont=sf,textfont=sf]{subfig}
\usepackage{textcomp}
\usepackage{stfloats}
\usepackage{url}
\usepackage{verbatim}
\usepackage{graphicx}
\usepackage{cite}
\usepackage{hyperref}
\usepackage{booktabs}
\graphicspath{{figures/}}
\DeclareGraphicsExtensions{{.pdf,.png}}
\usepackage{import}
\usepackage{adjustbox}
\usepackage[caption=false,font=normalsize,labelfont=sf,textfont=sf]{subfig}

\usepackage{multirow}
\usepackage{acronym}
\usepackage{siunitx}

\acrodef{emb}[EMB]{eye movement biometrics}
\acrodef{cnn}[CNN]{convolutional neural network}
\acrodef{resnet}[ResNet]{residual network}
\acrodef{densenet}[DenseNet]{densely connected convolutional network}
\acrodef{rnn}[RNN]{recurrent neural network}
\acrodef{del}[DEL]{Deep\-Eyedentification\-Live}
\acrodef{eky}[EKY]{Eye Know You}
\acrodef{far}[FAR]{false acceptance rate}
\acrodef{frr}[FRR]{false rejection rate}
\acrodef{eer}[EER]{equal error rate}
\acrodef{bn}[BN]{batch normalization}
\acrodef{relu}[ReLU]{rectified linear unit}

\begin{document}

\title{Eye Know You Too: A DenseNet Architecture for End-to-end Eye Movement Biometrics}

\author{Dillon Lohr, Oleg V Komogortsev 
\thanks{The authors are affiliated with the Department of Computer Science, Texas State University, San Marcos, TX 78666 USA. Author emails: \{djl70,ok11\}@txstate.edu.}
\thanks{Manuscript received April 19, 2021; revised August 16, 2021.}}

\markboth{Journal of \LaTeX\ Class Files,~Vol.~14, No.~8, August~2021}%
{Shell \MakeLowercase{\textit{et al.}}: A Sample Article Using IEEEtran.cls for IEEE Journals}

\IEEEpubid{0000--0000/00\$00.00~\copyright~2021 IEEE}

\maketitle

\import{sections/}{sec_01abstract}

\acresetall


\import{sections/}{sec_02intro}

\import{sections/}{sec_03litreview}

\import{sections/}{sec_04methods}

\import{sections/}{sec_05results}

\import{sections/}{sec_06discuss}

\import{sections/}{sec_07conclusion}

\section*{Acknowledgments}
This material is based upon work supported by the National Science Foundation Graduate Research Fellowship under Grant No. DGE-1144466.
The study was also funded by National Science Foundation grant CNS-1714623 to Dr. Komogortsev.
Any opinions, findings, and conclusions or recommendations expressed in this material are those of the author(s) and do not necessarily reflect the views of the National Science Foundation.

\bibliography{01main}
\bibliographystyle{IEEEtran}

\end{document}

%% file: sections/sec_01abstract.tex
\begin{abstract}
    \Ac{emb} is a relatively recent behavioral biometric modality that may have the potential to become the primary authentication method in virtual- and augmented-reality devices due to their emerging use of eye-tracking sensors to enable foveated rendering techniques.
    However, existing \ac{emb} models have yet to demonstrate levels of performance that would be acceptable for real-world use.
    Deep learning approaches to \ac{emb} have largely employed plain \acp{cnn}, but there have been many milestone improvements to convolutional architectures over the years including \acp{resnet} and \acp{densenet}.
    The present study employs a \ac{densenet} architecture for end-to-end \ac{emb} and compares the proposed model against the most relevant prior works.
    The proposed technique not only outperforms the previous state of the art, but is also the first to approach a level of authentication performance that would be acceptable for real-world use.
\end{abstract}

%% file: sections/sec_02intro.tex
\section{Introduction}
\IEEEPARstart{B}{iometrics} have become a part of everyday life due to the ubiquity of fingerprint and face recognition in smartphones.
Most biometric modalities can be separated into two categories: \textit{physical} and \textit{behavioral}.
Physical biometrics reflect the physical traits of a person, including face, fingerprint, iris, and retina.
Behavioral biometrics reflect a person's patterns of behavior, with some of the most commonly studied modalities being gait, signature, and voice.
Physical biometrics tend to be more distinctive and exhibit greater permanence, whereas behavioral biometrics tend to be less intrusive and more applicable for continuous authentication.
An overview of these common biometric modalities is given in~\cite{jain2004introduction}.

A more recent behavioral biometric modality is \ac{emb}~\cite{Kasprowski2004}.
Eye movements may be particularly spoof-resistant because the oculomotor system is controlled by a complex combination of neurological and physiological mechanisms, both voluntary and involuntary.
Eye movements also enable liveness detection~\cite{Komogortsev2015,makowski2021deepeyedentificationlive} and continuous authentication~\cite{Eberz2015,Eberz2019} and could easily be paired with other modalities like mouse dynamics~\cite{Kasprowski2018} or iris recognition~\cite{komogortsev2012multimodal} in a multimodal biometrics system.
Because eye movements have been shown to carry distinguishable information, studies have even begun to explore methods of deidentifying eye movement signals in an effort to preserve both the users' privacy and the utility of the eye movements as an input method~\cite{liebling2014privacy,liu2019differential,steil2019privacy,david-john2021privacy,li2021kaleido}.

There is an emerging use of eye-tracking sensors in both virtual-reality~(VR) and augmented-reality~(AR) devices (e.g., Vive Pro Eye~\cite{ViveProEye}, Magic Leap~1~\cite{ML1}, HoloLens~2~\cite{HL2}) in part to enable foveated rendering~\cite{guenter2012foveated} techniques which offer a significant reduction in overall power consumption without a noticeable impact to visual quality.
In addition to foveated rendering, eye tracking also enables various applications in these devices including user interactions, analytics, and various display technologies~\cite{stoffregen2022event}.
Because the hardware required for \ac{emb} is already included with these devices, and because \ac{emb} offers continuous authentication capabilities, \ac{emb} has the potential to become the primary security method for these devices~\cite{lohr2020vr}.
However, existing \ac{emb} models have yet to demonstrate levels of authentication performance that would be acceptable for real-world use, even when using eye-tracking signals with higher levels of signal quality than are available in current VR/AR devices.
Deep learning models for \ac{emb}~\cite{Jia2018,Abdelwahab2019,Jager2020,makowski2021deepeyedentificationlive,lohr2021eye} have largely been limited to plain \acfp{cnn} which, despite being capable of outperforming more traditional statistical approaches, do not take advantage of milestone developments over the years in the area of convolutional architectures, including \acfp{resnet}~\cite{he2015deep} and---the focus of the present study---\acfp{densenet}~\cite{huang2018densely}.

We propose a novel \ac{densenet}-based architecture for end-to-end \ac{emb}.
The model is trained and evaluated on the GazeBase~\cite{griffith2021gazebase} dataset which contains eye movement data from 322~subjects each recorded up to 18~times over a 37-month period.
Although the eye-tracking signals in GazeBase reflect a higher level of signal quality than is available in current VR/AR devices, it is important to first establish whether \ac{emb} is capable, even with high-quality data, of achieving a level of authentication performance that is acceptable for real-world use.
We primarily evaluate our model in the authentication scenario (also commonly called verification) using equal error rate~(EER), both because authentication performance has been shown to remain relatively stable regardless of population size~\cite{Friedman2020} and because behavioral biometric traits are generally not distinctive enough for large-scale identification~\cite{unar2014review}.
We also perform additional analyses that we have seen only a small portion of works do (e.g.,~\cite{lohr2021eye}), including assessing the permanence of the learned features and estimating the false rejection rate~(FRR) at a false acceptance rate~(FAR) of 1-in-10000 (abbreviated FRR @ FAR $10^{-4}$).

\IEEEpubidadjcol

The main contributions of the present study are:
\begin{itemize}
    \item A novel, highly parameter-efficient, \acs{densenet}-based architecture that achieves state-of-the-art \ac{emb} performance in the authentication scenario on high-quality data, namely 3.66\% EER when enrolling and authenticating with just 5~s of eye movements during a reading task.
    For perspective, 5~s is somewhat comparable to the time it takes to enter a 4-digit pin or to calibrate an eye-tracking device.
    \item The first to show that, using 30~s of eye movements during a reading task, it is possible to achieve an estimated 5\% FRR @ FAR $10^{-4}$ which approaches a level of authentication performance that would be acceptable for real-world use.
    \item The first to report significantly better-than-chance FRR @ FAR $10^{-4}$ with 60~s of eye movements at artificially degraded sampling rates as low as 50~Hz, suggesting that \ac{emb} has the potential to become suitable for deployment at the sampling rates present in existing VR/AR devices.
    \item The first application of a more modern convolutional architecture for \ac{emb}.
\end{itemize}

\begin{figure*}[!t]
    \centering
    \includegraphics[width=\linewidth]{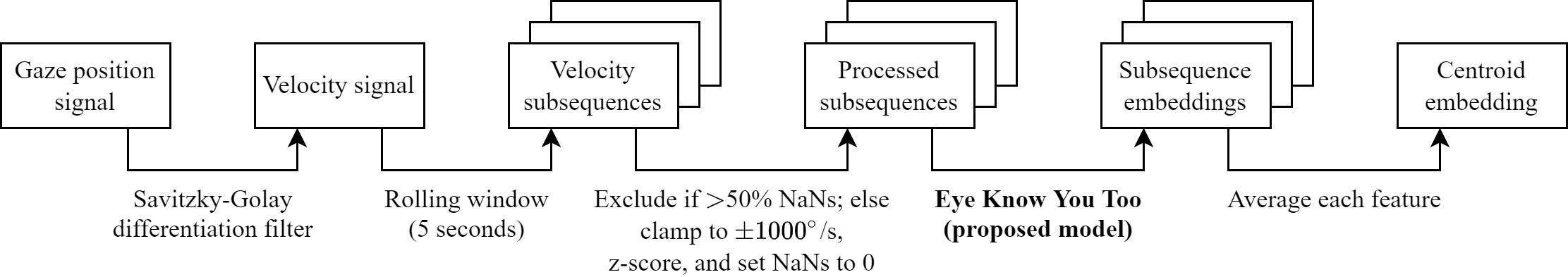}
    \caption{Overview of the process for embedding an eye-tracking signal using the proposed methodology. We primarily focus on the case where only the first 5-second window is embedded, but we explore aggregating embeddings across windows in \S~\ref{sec:results-window}.}
    \label{fig:overview}
\end{figure*}

%% file: sections/sec_03litreview.tex
\section{Prior Work}
\label{sec:prior-work}
\subsection{\Acfp{cnn}}
\noindent Since the seminal works of AlexNet~\cite{alex2012imagenet} and VGGNet~\cite{simonyan2015deep}, \acp{cnn} have quickly become some of the most popular types of neural networks for image processing tasks.
Such architectures also started being employed in time series domains like eye movement event classification~\cite{elmadjian2021tcn} and audio synthesis~\cite{oord2016wavenet}.
In time series domains such as these, varieties of recurrent neural networks~(RNNs)~\cite{hochreiter1997lstm,cho2014learning} were once the most common, but \acp{cnn} have empirically shown to be capable of similar-or-better performance while also being much faster to train~\cite{lea2016temporal}.

Several pivotal architectural improvements have been made to \acp{cnn} since their infancy.
We focus on two such improvements: \acp{resnet}~\cite{he2015deep} and \acp{densenet}~\cite{huang2018densely}.
\Acp{resnet}~\cite{he2015deep} introduce so-called ``skip connections'' that combine the output of each convolutional block with its input via summation.
These skip connections improve gradient flow through the network, enabling the training of significantly deeper networks than was previously possible.
\Acp{densenet}~\cite{huang2018densely} include similar skip connections between each convolutional block and \textit{all} subsequent blocks, using channel-wise concatenation instead of summation to facilitate even better information flow than \acp{resnet}.
One study visualizing loss landscapes~\cite{li2018visualizing} showed that \acp{densenet} have much smoother loss landscapes than \acp{resnet} which may lead to increased ease of convergence during training.

We acknowledge that there are more recent convolutional architectures than \ac{densenet} that claim better performance on image processing tasks (e.g., ResNeXT~\cite{xie2017aggregated}, DSNet~\cite{zhang2020resnet}, EfficientNet~\cite{tan2020efficientnet}, EfficientNetV2~\cite{tan2021efficientnetv2}), not to mention the various transformer~\cite{vaswani2017attention,dosovitskiy2021vit} architectures that have seen success in domains including natural language processing~(NLP) and image classification.
Rather than using one of these more cutting-edge architectures, we base our architecture on \ac{densenet} because of its parameter efficiency and relative simplicity.
We find that this architecture is able to achieve state-of-the-art performance in the \ac{emb} domain.

\subsection{\Acf{emb}}
\noindent \Ac{emb} has been studied extensively since the introduction of the modality in 2004~\cite{Kasprowski2004}.
Most earlier works in the field~\cite{holland2011biometric,george2016score,Friedman2017,lohr2020metric,porta2022gaze} require explicit classification of eye movement signals into physiologically-grounded events, from which hand-crafted features are extracted and fed into statistical or machine learning models.
The state-of-the-art statistical model is the approach by Friedman et al.~\cite{Friedman2017} which centers around the use of principal component analysis~(PCA) and the intraclass correlation coefficient~(ICC).

Since the recent introduction of deep learning to the field of \ac{emb}~\cite{Jia2018, Abdelwahab2019}, end-to-end deep learning approaches have become more common~\cite{Jia2018,Abdelwahab2019,Jager2020,makowski2021deepeyedentificationlive,lohr2021eye}.
The current state-of-the-art model is \ac{del}~\cite{makowski2021deepeyedentificationlive} which utilizes two convolutional subnets that separately focus on ``fast'' (e.g., saccadic) and ``slow'' (e.g., fixational) eye movements.
Another recent model, \ac{eky}~\cite{lohr2021eye}, uses exponentially dilated convolutions to achieve reasonable biometric authentication performance with a relatively small ($\sim$475K learnable parameters) network architecture.

Both \ac{del} and \ac{eky} employ plain \ac{cnn} architectures that do not take advantage of the improvements made to \acp{cnn} over the years.
The present study improves upon the previous state of the art by using a more modern \ac{densenet}-based architecture to simultaneously increase expressive power and reduce parameter count.

%% file: sections/sec_04methods.tex
\section{Network Architecture}
\label{sec:network-arch}
\begin{figure*}[!t]
    \centering
    \includegraphics[width=\textwidth]{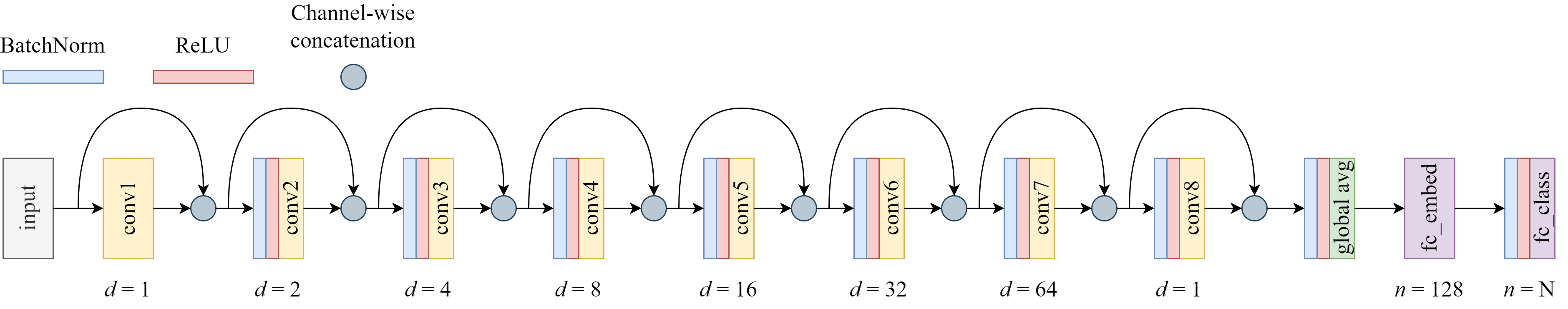}
    \caption{The proposed pre-activation \ac{densenet}-based network architecture, including the optional classification layer. Each convolution layer has kernel size $k=3$, stride $s=1$, and dilation rate $d$ that varies by layer. Each convolution layer outputs 32~feature maps that are concatenated with the previous feature maps before being fed into the next convolution layer.}
    \label{fig:network-arch}
\end{figure*}

\noindent The proposed network architecture, which we call Eye Know You Too~(EKYT), is visualized in Fig.~\ref{fig:network-arch}.
The network performs a mapping $f: \mathbb{R}^{C \times T} \to \mathbb{R}^{128}$, where $C$ is the number of input channels, $T$ is the input sequence length, and the output is a 128-dimensional embedding.
It begins with a single dense block of 8~one-dimensional convolution layers, where the feature maps produced by each convolution layer are concatenated with all previous feature maps before being fed into the next convolution layer.
The final set of concatenated feature maps is then sent through a global average pooling layer, flattened, and then fed into a fully-connected layer to produce a 128-dimensional embedding of the input sequence.
When classification is required (e.g., for cross-entropy loss), an additional fully-connected layer is appended after the embedding layer that outputs class logits.

All convolution layers (except the first), the global average pooling layer, and the optional classification layer are all preceded by \ac{bn}~\cite{ioffe2015batchnorm} and the \ac{relu}~\cite{nair2010relu} activation function (called a ``pre-activation'' architecture).
We use a ``growth rate'' of~32, meaning each convolution layer outputs 32~feature maps to be concatenated with the previous feature maps.
Because we use \ac{bn}, there is no need for the convolution layers to learn an additive bias.

The convolution layers (labeled $n = 1, \dotsc, 8$) use constant kernel size $k=3$ and stride $s=1$, an exponentially increasing dilation rate $d_{n} = 2^{(n - 1) \bmod 7}$, and enough zero padding $p_{n}=d_{n}$ on both sides of the input to preserve the length along the feature dimension.
The use of exponentially dilated convolutions produces an exponential growth of the receptive field of the network with only a linear increase in the number of learnable parameters.
In general, assuming $s=1$, the receptive field of layer $n$, denoted $r_{n}$, is given by 
\begin{equation}
    r_{n} = 1 + \sum_{i=1}^{n}{d_{n}(k_{n}-1)}.
\end{equation}
The final convolution layer of our network has a (maximum) receptive field of $r_8 = 257$ time steps from the input.

Excluding the optional classification layer, our proposed architecture has $\sim$123K learnable parameters for $C=2$ and any $T$.
Weights are initialized in the following manner.
Each convolutional layer uses He initialization~\cite{he2015delving} with a normal distribution and learns no additive bias.
Each \ac{bn} layer is initialized with a weight of~1 and a bias of~0.
Each fully-connected layer is initialized with a bias of~0, and weights are initialized using the default method of PyTorch~1.10.0.

In preliminary experiments, we experienced overfitting on the train set relative to the validation set when increasing the depth of the network and/or adding additional dense blocks (each separated by a transition block to optionally reduce the size of the channel dimension).
Specialized dropout techniques have been proposed for \ac{densenet} architectures to resolve such overfitting problems~\cite{wan2019reconciling}; but in the interest of keeping our network small, we did not pursue such techniques.
We also experienced worse performance when using global max pooling instead of global average pooling.
We found no noticeable difference in performance when swapping the order of \ac{bn} and \ac{relu}, nor when using a ``post-activation'' architecture (i.e., applying \ac{bn} and \ac{relu} after each convolution before channel-wise concatenation).
Though, we note that pre-activation \acs{densenet} (and \acs{resnet}) architectures generally produce lower errors than their post-activation counterparts~\cite{he2016identity}.

\section{Methodology}
\label{sec:methodology}

\subsection{Hardware \& software}
\noindent All models are trained inside Docker containers on a Lambda Labs workstation.
The workstation is equipped with quad NVIDIA GeForce RTX A5000 GPUs (24~GB VRAM), an AMD Ryzen Threadripper PRO 3975WX CPU @ 3.5~GHz (32~cores), and 256~GB RAM.
Each Docker container runs Ubuntu~18.04 with the most notable packages being Python~3.7.11, PyTorch~\cite{paszke2019pytorch}~1.10.0, and PyTorch Metric Learning~(PML)~\cite{Musgrave2020a}~0.9.99.
PyTorch Lightning~\cite{falcon2019pl}~1.5.0 is used to accelerate development.
Experiments are logged using Weights \& Biases~\cite{biewald2020wandb}~0.12.1.
For visualizing the embedding space, we employ umap-learn~\cite{mcinnes2018umap-software}~0.5.1.

Our full source code and trained models are available on the Texas State Digital Collections Repository at [TO BE
PUBLISHED AFTER ACCEPTANCE].

\subsection{Dataset}
\noindent We use the GazeBase~\cite{griffith2021gazebase} dataset consisting of 322~college-aged participants, each recorded monocularly (left eye only) at 1000~Hz using an EyeLink~1000 eye tracker.
Nine rounds of recordings (R1--9) were captured over a period of 37~months.
Each subsequent round comprises a subset of participants from the preceding round (with one exception, participant~76, who was not present in R3 but returned for R4--5), with only 14~of the initial 322~participants present across all 9~rounds.
Each round consists of 2~recording sessions separated by approximately 30~minutes.
During each recording session, participants performed a battery of 7~tasks: random saccades~(RAN), horizontal saccades~(HSS), fixation~(FXS), an interactive ball-popping game~(BLG), reading~(TEX), and two video-viewing tasks~(VD1 and VD2).
More details about each task can be found in the dataset's paper~\cite{griffith2021gazebase}.

We create class-disjoint train and test sets by assigning the 59~participants present during R6 to the test set and the remaining 263~participants to the train set.
In this way, the test set---which comprises nearly 50\% of all recordings in GazeBase---can be used to assess the generalizability of our model both to out-of-sample participants and to longer test-retest intervals than are present during training.
The train set is further partitioned into 4~class-disjoint folds in a way that balances the number of participants and recordings between folds as well as possible (the fold assignment algorithm we use is described in~\cite{lohr2021eye}).
These 4~folds are used for 4-fold cross-validation, where 1~fold acts as the validation set and the remaining folds act as the train set.
We exclude the BLG task from the train and validation sets due to the large variability in its duration relative to the other tasks, but we include it in the test set to enable an assessment of our model on an out-of-sample task.

We used 4-fold cross-validation to manually tweak our network architecture and determine the final training parameters.
Only at the very end of our experiments did we use the test set to get a final, unbiased estimate of our model's performance.

\subsection{Data preprocessing}
\noindent We start with a sequence of $T_{\text{record}}$ tuples $\left(t^{(i)},\, x^{(i)},\, y^{(i)}\right)$, $i = 1,\dotsc,T_{\text{record}}$, where $t^{(i)}$ is the time stamp (s) and $x^{(i)}$, $y^{(i)}$ are the horizontal and vertical components of the monocular (left eye) gaze position ($^{\circ}$).
Next, we estimate the first derivative (i.e., velocity in $^{\circ}$/s) of the horizontal and vertical channels using a Savitzky-Golay~\cite{savitzky1964sgolay} differentiation filter with order~2 and window size~7, inspired by~\cite{Friedman2017}.

Each recording is then split into non-overlapping windows of 5~s ($T=5000$~time steps) using a rolling window.
Excess time steps at the end of a recording that would form only a partial window are discarded.
Although previous studies like \ac{del}~\cite{makowski2021deepeyedentificationlive} and \ac{eky}~\cite{lohr2021eye} use input sizes of around 1~s, we found in our experiments that our model performs better when using a larger input size of 5~s than when it has to learn from isolated samples of 1~s.
We believe this is because the model can take advantage of longer-term patterns when given longer sequences.
We note that 5~s is somewhat comparable to the amount of time it takes to enter a 4-digit pin or to calibrate an eye-tracking device.

Velocities are clamped between $\pm 1000^{\circ}$/s to limit the influence of noise.
They are then z-score transformed using a single mean and standard deviation determined across both channels in the train set.
Finally, any NaN values are replaced with 0 after z-scoring. 

We observed during our experiments that estimating velocity with a Savitzky-Golay differentiation filter and scaling with a z-score transformation led to marginal improvements in performance metrics compared to the approach of~\cite{makowski2021deepeyedentificationlive}, wherein velocity is computed with the two-point central difference method and velocities are transformed using the ``fast'' and ``slow'' transformations proposed in~\cite{Jager2020}.

\subsection{Training}
\noindent Input samples consist of windows of $T=5000$~time steps and $C=2$ channels: horizontal and vertical velocity.
Following~\cite{lohr2021eye}, we primarily use multi-similarity~(MS)~\cite{Wang2019} loss to train our model.
MS loss encourages the learned embedding space to be well-clustered, meaning a sample from one class is closer to other samples from the same class than to samples from different classes. 
However, we observed during our experiments that a weighted sum of MS loss (using the output of the embedding layer) and categorical cross-entropy~(CE) loss (using the output of the classification layer) led to marginal improvements in performance metrics compared to using MS loss alone.

Therefore, our loss function $L$ is given by the following equations:
\begin{align}
\begin{split}
    L_{MS} = \frac{1}{m} \sum_{i=1}^{m} &\left( \frac{1}{\alpha} \log \left( 1 + \sum_{k \in P_i}{ \exp{\left(-\alpha\left(S_{ik} - \lambda\right)\right)} } \right) \right. \\
    &+ \left. \frac{1}{\beta} \log \left( 1 + \sum_{k \in N_i}{ \exp{\left(\beta\left(S_{ik} - \lambda\right)\right)} } \right) \!\right),
\end{split}
\end{align}
\begin{equation}
L_{CE} = \frac{1}{m} \sum_{i=1}^{m} -\log{\left( \frac{\exp{\left(x_{i,y_i}\right)}}{\sum_{j=1}^{N} \exp{\left(x_{i,j}\right)}} \right)},
\end{equation}
\begin{equation}
L = w_{MS} L_{MS} + w_{CE} L_{CE},
\end{equation}
where $w_{MS} = 1.0$ and $w_{CE} = 0.1$ are the weights for the respective loss functions; $m=256$ is the size of each minibatch; $\alpha=2.0$, $\beta=50.0$, and $\lambda=0.5$ are hyperparameters for MS loss; $P_i$ and $N_i$ are the sets of indices of the mined positive and negative pairs for each anchor sample $\mathbf{x}_i$; $S_{ik}$ is the cosine similarity between the pair of samples $\{\mathbf{x}_i,\, \mathbf{x}_k\}$; $N$ is the number of classes (either 197 or 198) in the train set; $x_{i,j}$ is the predicted logit for sample $i$ and class label $j$; and $y_i$ is the target class label for sample $i$.
The above formulation for MS loss implicitly includes an online pair miner with an additional hyperparameter $\varepsilon=0.1$.
More details about MS loss can be found in~\cite{Wang2019}.

Each minibatch consists of 256~samples constructed in the following manner.
First, 16~unique subjects are selected at random from the train set.
Next, 16~windows are randomly selected without replacement for each of the selected subjects.
These windows could be selected from any of the rounds, sessions, and tasks that each subject was present for in the train set.
This results in 256~windows per minibatch.
Each training ``epoch'' iterates over as many minibatches as needed until a number of windows, equivalent to the total number of unique windows in the train set, has been sampled.
Note that because of the nature of this minibatch construction method, windows from earlier rounds may be over-represented~\cite{lohr2021eye}, and not every window from the train set may be included in any given epoch.

One downside of using input windows of 5~s is that it greatly reduces the number of samples available for training compared to when using a smaller input window like 1~s.
This is a problem because deep learning methodologies generally perform better when trained on larger datasets.
We are able to mitigate this issue by training on all tasks (except BLG) simultaneously instead of training on a single task.
Additionally, the use of varied tasks encourages learned features to be informative for all types of eye movements (particularly fixations, saccades, and smooth pursuits) and enables a single model to be applied on multiple tasks, instead of necessitating a separate model for different tasks as most prior works do.

\begin{figure}
    \centering
    \subfloat[][]{
        \label{fig:lr}
        \includegraphics[width=0.47\linewidth]{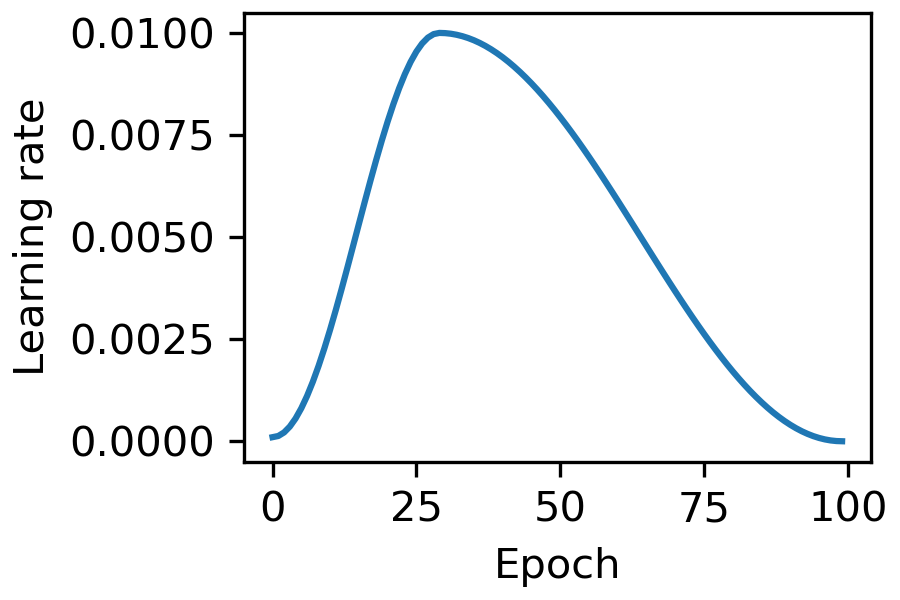}
    }
    \subfloat[][]{
        \label{fig:mapr}
        \includegraphics[width=0.47\linewidth]{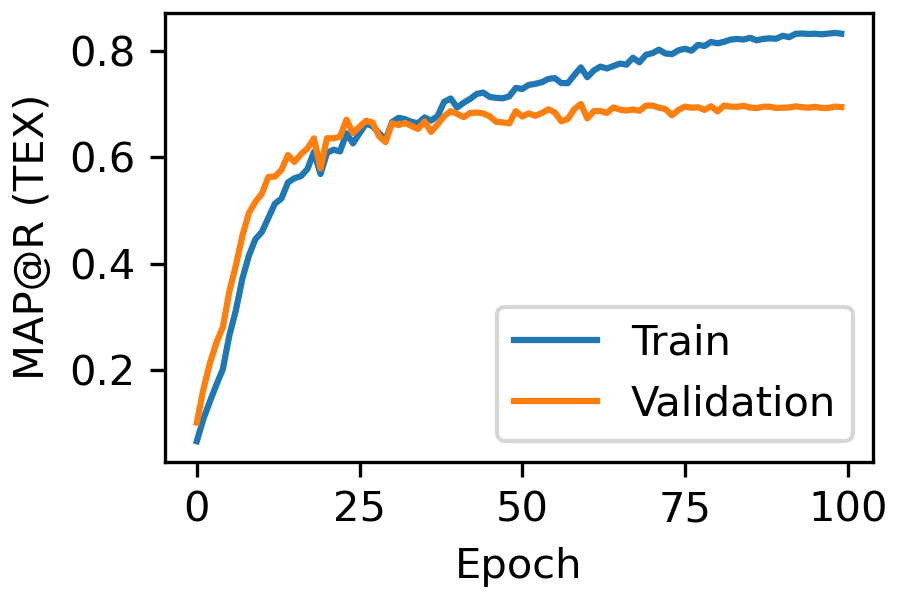}
    }
    \caption{(A) A visualization of the learning rate schedule used during training. (B) Progression of MAP@R (measured on TEX only) throughout training.}
    \label{fig:lr-mapr}
\end{figure}

We employ the Adam~\cite{kingma2014adam} optimizer with a one-cycle cosine annealing learning rate scheduler~\cite{smith2018superconvergence} (visualized in Fig.~\ref{fig:lr}).
The learning rate starts at $10^{-4}$, gradually increases to a maximum of $10^{-2}$ over the first 30~epochs, and then gradually decreases to a minimum of $10^{-7}$ over the next 70~epochs.
We found that, compared to using a fixed learning rate throughout training, this learning rate schedule both accelerated the training process and led to higher levels of performance.
Training lasts for a fixed duration of 100~epochs, and the final weights of the model are saved.

We note that there is no feedback from the validation set when training in this way (in contrast to when early stopping is employed, for example).
However, the validation set was used while manually tweaking the proposed architecture and training paradigm.
The final architecture was chosen as the one that maximized Mean Average Precision at R~(MAP@R)~\cite{Musgrave2020} on the validation set (using embeddings from TEX only).
MAP@R is a clustering metric that we believe is more informative than \acs{eer} for model selection.
We visualize the progression of MAP@R throughout training in Fig.~\ref{fig:mapr} to provide some insight into the values we achieve; but because it is not directly related to biometric authentication, we do not report MAP@R in our results.

\subsection{Evaluation}
\noindent Although the model is trained on samples from all tasks (except BLG), we primarily evaluate the model on only TEX due to the prevalent usage of reading data in the \ac{emb} literature.

We primarily evaluate the model using \acf{eer}, which is the point where \ac{frr} is equal to \ac{far}.
Measuring \ac{eer} requires a set of data used for enrollment and a separate set of data for authentication (also commonly called verification).
The enrollment set is formed using the first window (5~s) of the session~1 TEX task from R1 for each subject in the test set.
The authentication set is formed using the first window (5~s) of the session~2 TEX task from R1 for each subject in the test set.

To ensure a minimal level of sample fidelity at evaluation time, we discard windows with more than 50\% NaNs.
Subjects are effectively excluded from the enrollment or authentication sets if they have no valid windows in the respective set.

For each window in the enrollment and authentication sets, we compute the 128-dimensional embeddings with each of the 4~models trained with 4-fold cross-validation.
We then concatenate these embeddings to form a single, 512-dimensional embedding for each window, effectively treating the 4~models as a single ensemble model.
We compute all pairwise cosine similarities between the embeddings in the enrollment set and those in the authentication set.
The resulting similarity scores are fed into a receiver operating characteristic~(ROC) curve to measure \ac{eer}.

Given the resolution of a particular ROC curve, there may not be a similarity threshold where \ac{frr} and \ac{far} are exactly equal.
In such cases, the \ac{eer} needs to be estimated, and there are several ways this estimation can be done.
The method we use is to linearly interpolate between the points on the ROC curve to estimate the point where \ac{frr} and \ac{far} would be equal.

In addition to the primary evaluation setting described above, we can also change different parameters to evaluate our model under various conditions.
We will explore our model's performance on different tasks, across longer test-retest intervals, and using increasing amounts of data for enrollment and authentication.
We will also examine how well our network adapts to data with lower sampling rates.
These additional analyses will be described later.

%% file: sections/sec_05results.tex
\section{Results}
\label{sec:results}
\noindent Unless otherwise specified, presented results are measured on the held-out test set using an ensemble model evaluated under the primary evaluation setting, meaning we enroll and authenticate with 5~s of 1000~Hz data from R1 TEX.

\begin{figure}[!t]
    \centering
    \subfloat[][Similarity distributions]{
        \label{fig:hist}
        \includegraphics[width=0.47\linewidth]{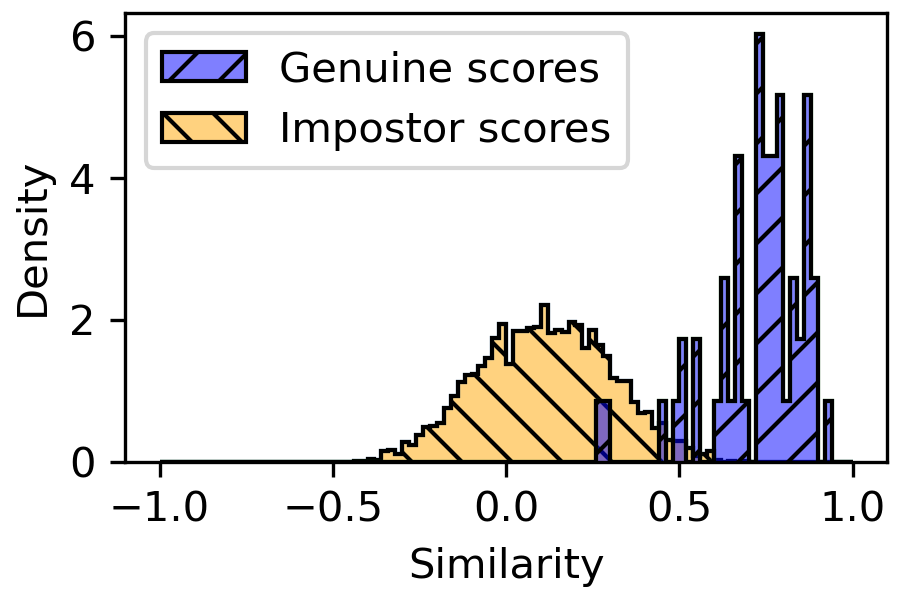}
    }
    \subfloat[][ROC curve]{
        \label{fig:roc}
        \includegraphics[width=0.47\linewidth]{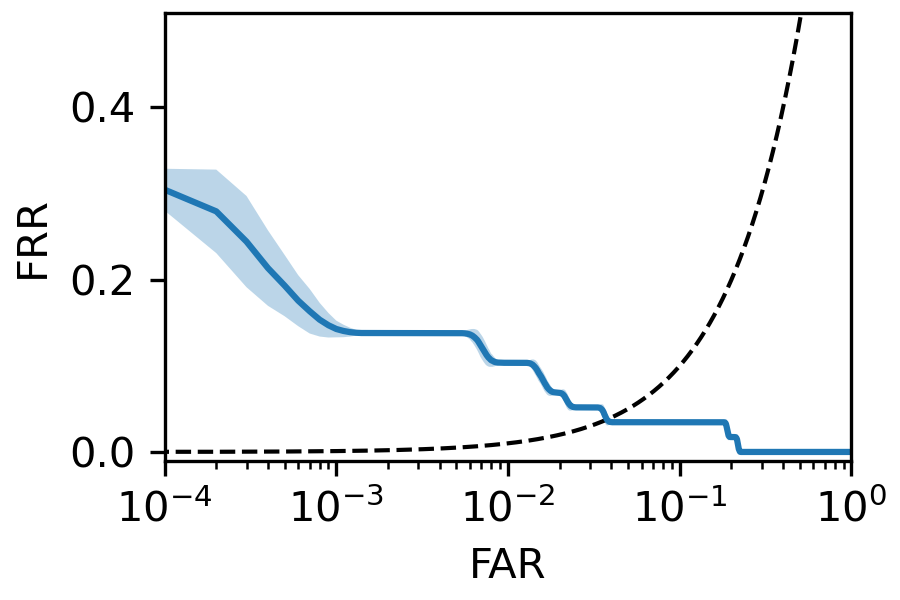}
    }
    \caption{Qualitative results for the primary evaluation setting: 5~s of R1 TEX. (A) Genuine and impostor similarity score distributions. (B) ROC curve for bootstrapped similarity score distributions (see \S~\ref{sec:results-bootstrap} for an explanation of how the bootstrapped distributions are made). The dashed black line shows where FRR and FAR are equal. The blue line is the mean ROC curve across 1000~bootstrapped distributions, and the shaded region represents $\pm$1~standard deviation around the mean.}
    \label{fig:hist-roc}
\end{figure}

Results on the test set for our primary evaluation setting are presented in the first row of Table~\ref{tab:main-results-noexclude-train-drop50}.
The other results in that table are described in the upcoming subsections.
Fig.~\ref{fig:hist-roc} shows the similarity score distributions and ROC~curve under the primary evaluation setting.

\begin{figure}[!t]
    \centering
    \subfloat[][TEX only]{
        \label{fig:densmap-tex}
        \includegraphics[width=0.47\linewidth]{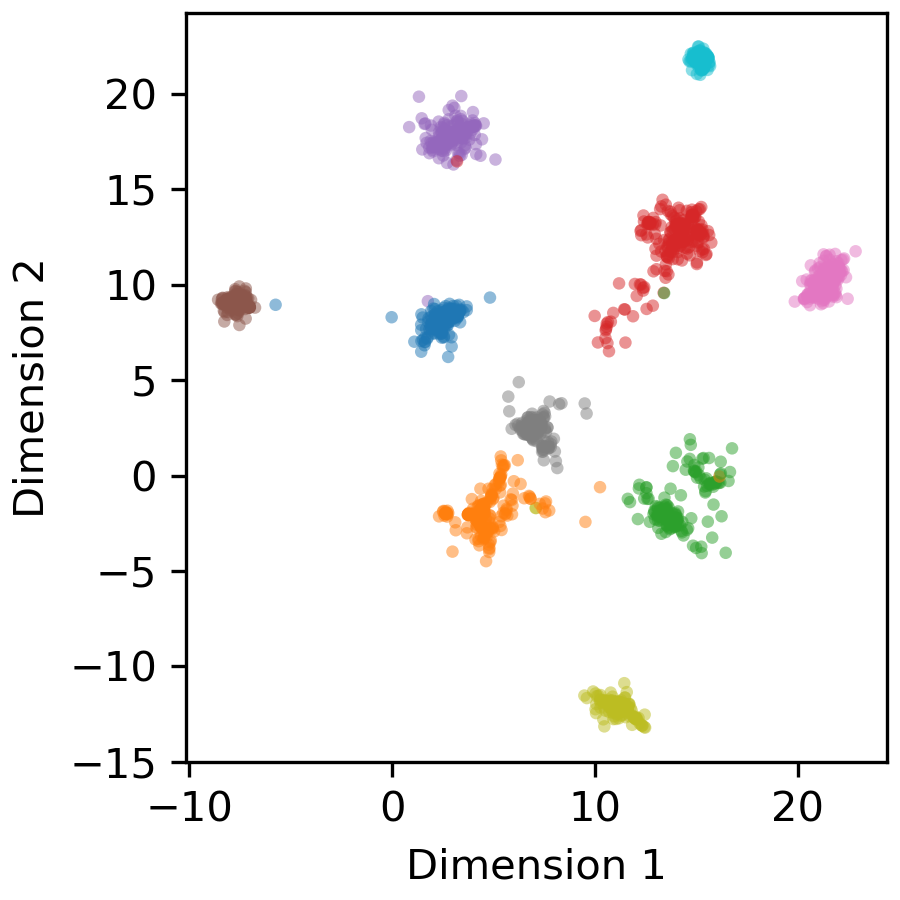}
    }
    \subfloat[][All tasks $+$ BLG]{
        \label{fig:densmap-all}
        \includegraphics[width=0.47\linewidth]{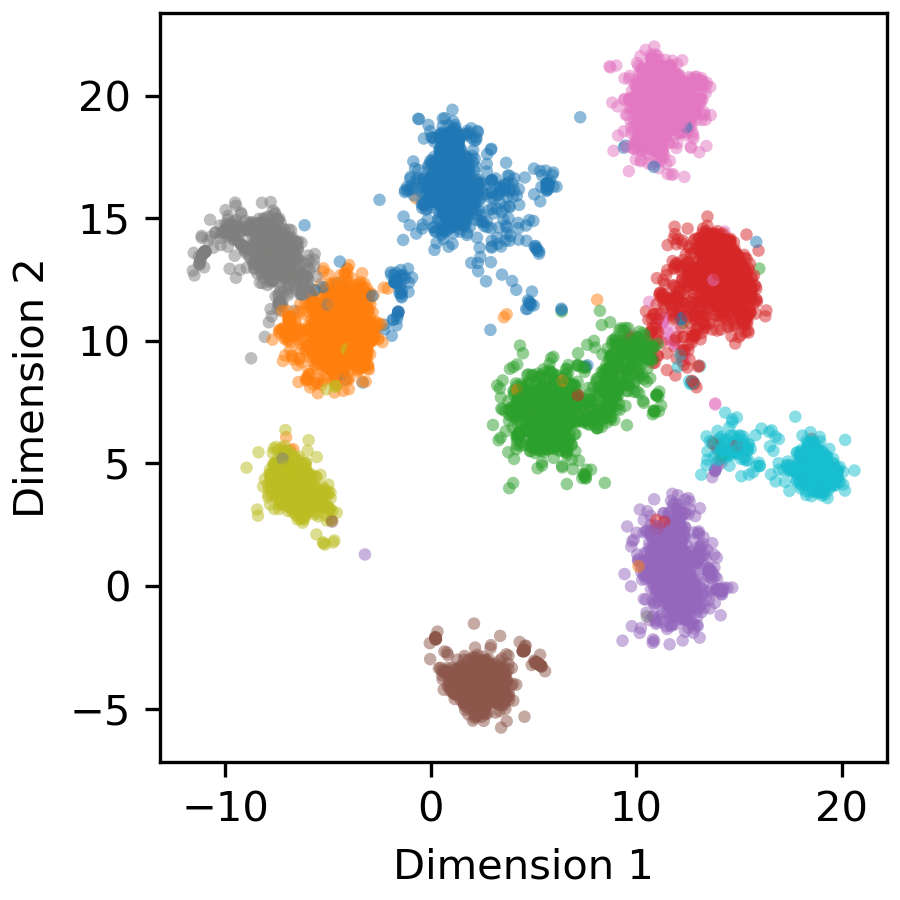}
    }
    \caption{DensMAP~\cite{narayan2020densmap} visualizations of the embedding space for 10~subjects present across all rounds. All embeddings of valid ($\leq$50\% NaNs) windows across all rounds R1--9 and both sessions are plotted together. A different mapping is fit for each plot. (A) Embeddings from only the TEX task. (B) Embeddings from all tasks (including BLG). We use umap-learn~\cite{mcinnes2018umap-software} parameters metric=cosine, n\_neighbors=30, min\_dist=0.1, and densmap=True.}
    \label{fig:densmap}
\end{figure}

To visualize the embedding space, DensMAP~\cite{narayan2020densmap} is used to create a low-dimensional representation of the embedding space in a way that attempts to globally and locally preserve structure and density.
A subset of the embedding space is visualized in Fig.~\ref{fig:densmap}.

Although we focus on the authentication scenario, it is worth briefly mentioning for completeness how the model performs in the identification scenario.
We employ the rank-1 identification rate which measures how often the correct identities have the highest similarity score between the enrollment and authentication sets.
Under the primary evaluation setting, after removing any authentication subjects who are not present in the enrollment set, rank-1 identification rate is 91.38\% (53~of 58~subjects are correctly identified).

\begin{table}[!t]
    \centering
    \caption{Biometric authentication results for various evaluation settings using a single ensemble of models trained with 4-fold cross-validation. Duration is given as $T \times n$, where $T$ is the length of each sample and $n$ is the number of samples. P and N are the numbers of positive and negative pairs, respectively.}
    \label{tab:main-results-noexclude-train-drop50}
    \begin{tabular}{@{} c c c l S[table-format=1.2] S[table-format=2.0] S[table-format=4.0] @{}}
        \toprule
        {\parbox{1cm}{\centering Effect}} & {Duration (s)} & {Round} & {\centering Task} & {EER (\%)} & {P} & {N} \\
        \midrule
        {-} & $5\times1$ & R1 & TEX & 3.66 & 58 & 3364 \\
        \midrule
        \midrule
        \multirow{6}{*}{\rotatebox{90}{\parbox{1.8cm}{\centering Task\\(\S~\ref{sec:results-task})}}} & \multirow{6}{*}{$5\times1$} & \multirow{6}{*}{R1} & HSS & 5.08 & 59 & 3422 \\
        {} & {} & {} & RAN & 5.08 & 59 & 3422 \\
        {} & {} & {} & FXS & 9.38 & 59 & 3422 \\
        {} & {} & {} & VD1 & 5.45 & 55 & 3135 \\
        {} & {} & {} & VD2 & 3.39 & 59 & 3422 \\
        {} & {} & {} & BLG* & 5.49 & 59 & 3422 \\
        \midrule
        \multirow{8}{*}{\rotatebox{90}{\parbox{2.4cm}{\centering Test-retest interval\\(\S~\ref{sec:results-round})}}} & \multirow{8}{*}{$5\times1$} & R2 & \multirow{8}{*}{TEX} & 8.62 & 58 & 3364 \\
        {} & {} & R3 & {} & 7.43 & 58 & 3364 \\
        {} & {} & R4 & {} & 8.71 & 58 & 3364 \\
        {} & {} & R5 & {} & 7.14 & 58 & 3306 \\
        {} & {} & R6 & {} & 6.09 & 58 & 3364 \\
        {} & {} & R7 & {} & 8.52 & 34 & 1996 \\
        {} & {} & R8 & {} & 8.89 & 30 & 1710 \\
        {} & {} & R9 & {} & 7.69 & 13 & 799 \\
        \midrule
        \multirow{11}{*}{\rotatebox{90}{\parbox{3.3cm}{\centering Duration\\(\S~\ref{sec:results-window})}}} & $5\times\hphantom{1}2$ & \multirow{11}{*}{R1} & \multirow{11}{*}{TEX} & 2.23 & 58 & 3364 \\
        {} & $5\times\hphantom{1}3$ & {} & {} & 0.76 & 59 & 3422 \\
        {} & $5\times\hphantom{1}4$ & {} & {} & 0.38 & 59 & 3422 \\
        {} & $5\times\hphantom{1}5$ & {} & {} & 0.58 & 59 & 3422 \\
        {} & $5\times\hphantom{1}6$ & {} & {} & 0.56 & 59 & 3422 \\
        {} & $5\times\hphantom{1}7$ & {} & {} & 0.58 & 59 & 3422 \\
        {} & $5\times\hphantom{1}8$ & {} & {} & 0.56 & 59 & 3422 \\
        {} & $5\times\hphantom{1}9$ & {} & {} & 0.56 & 59 & 3422 \\
        {} & $5\times10$ & {} & {} & 0.50 & 59 & 3422 \\
        {} & $5\times11$ & {} & {} & 0.41 & 59 & 3422 \\
        {} & $5\times12$ & {} & {} & 0.58 & 59 & 3422 \\
        \bottomrule
        \multicolumn{7}{l}{*~BLG was not included in the train or validation sets.}
    \end{tabular}
\end{table}

\subsection{Effect of task on authentication accuracy}
\label{sec:results-task}
\noindent For this analysis, we replace TEX with one of the other tasks during evaluation and repeat for each task.
Note that we evaluate our single ensemble model across all tasks; we do not train a separate model for each task.
Results are presented in Table~\ref{tab:main-results-noexclude-train-drop50} in the ``Task'' effect group.
To assess our model's performance on an out-of-sample task, we also include results for BLG.

\subsection{Effect of test-retest interval on authentication accuracy}
\label{sec:results-round}
\noindent For this analysis, we continue using the first session of R1 for the enrollment set, but for the authentication set we use the second session of one of the later rounds (R2--9) to assess how robust our model is to template aging after as many as 37~months.
Results are presented in Table~\ref{tab:main-results-noexclude-train-drop50} in the ``Test-retest interval'' effect group.

\subsection{Effect of recording duration on authentication accuracy}
\label{sec:results-window}
\noindent For this analysis, instead of limiting ourselves to the first 5-second window of a recording, we aggregate embeddings across the first $n$ windows to form a new, centroid embedding.
Since the model is trained to create a well-clustered embedding space, averaging multiple embeddings for a given class should lead to a better estimate of that class's central tendency in the embedding space.
Results are presented in Table~\ref{tab:main-results-noexclude-train-drop50} in the ``Duration'' effect group.

\subsection{Effect of sampling rate on authentication accuracy}

\noindent For this analysis, instead of using 1000~Hz data, we downsample each recording to different target sampling rates using an anti-aliasing filter (SciPy's~\cite{SciPy2020} \texttt{decimate} function) to assess how robust our network architecture is to lower sampling rates.
The targeted sampling rates are the same as those in~\cite{lohr2021eye}: 500, 250, 125, 50, and 31.25~Hz.
Input size is reduced by the same integer factor as the sampling rate and then truncated to remove any fractional components.
For example, at 31.25~Hz (a downsample factor of~32), the input size becomes $\lfloor \frac{5000}{32} \rfloor = 156$~time steps.

Since our network architecture contains a global pooling layer prior to the fully-connected layer(s), the network can be applied to time series of any length without any modifications.
But, because features learned at one sampling rate would not likely translate well to different sampling rates, we opted to train a new ensemble of 4-fold cross-validated models for each degraded sampling rate to have the best chance at extracting meaningful information at each sampling rate.

We do not adjust the Savitzky-Golay differentiation filter parameters for the lower sampling rates.
It is also worth noting that the (maximum) receptive field of our network, 257~time steps, is larger than the input sizes at 50~and 31.25~Hz.

\begin{table}[!t]
    \centering
    \caption{Biometric authentication results at degraded sampling rates. A different ensemble of models is trained for each sampling rate.}
    \label{tab:results-degraded}
    \begin{tabular}{@{} S[table-format=3.2] S[table-format=2.2] S[table-format=2.0] S[table-format=4.0] @{}}
        \toprule
        {Sampling rate (Hz)} & {EER (\%)} & {P} & {N} \\
        \midrule
        500  & 5.66 & 53 & 3079 \\
        250 & 6.20 & 53 & 3079 \\
        125 & 8.77 & 57 & 3306 \\
        50 & 15.52 & 58 & 3364 \\
        31.25 & 23.37 & 58 & 3364 \\
        \bottomrule
    \end{tabular}
\end{table}

\subsection{Estimating FRR @ FAR $10^{-4}$}
\label{sec:results-bootstrap}
\noindent The ultimate goal of \ac{emb} is to enable the use of eye movements for biometric authentication in real-world settings.
It is important to consider how \ac{emb} compares to existing security methods, because if it cannot outperform such methods then wide adoption would be unlikely.
Like~\cite{lohr2021eye}, we use the 4-digit (10-key) pin as a representative for existing security methods, because it is one of the most common security methods in everyday life as both a primary and secondary security measure.
The 4-digit pin effectively has a \ac{far} of 1-in-10000, assuming each of the $10^4$ combinations of 4-digit 10-key pins is equally likely to be chosen by enrolled users.

For this analysis, to mimic the level of security afforded by a 4-digit pin, we provide estimates of FRR when FAR is fixed at $10^{-4}$ (abbreviated FRR @ FAR $10^{-4}$).
Directly measuring FRR @ FAR $10^{-4}$ requires at least $\text{N}=10000$~impostor similarity scores, but we are limited to a maximum of 3422.
Therefore, to enable the estimation of FRR @ FAR $10^{-4}$, we use bootstrapping (i.e., repeated random sampling with replacement) to resample our empirical genuine and impostor similarity score distributions to form new distributions with $\text{P}=20000$ and $\text{N}=20000$ scores.
We repeat bootstrapping 1000~times and report the mean and standard deviation of the performance across those 1000~bootstrapped distributions.
The FIDO Biometrics Requirements~\cite{FIDO2020} suggest that a biometric system should have no higher than 3--5\% FRR @ FAR $10^{-4}$, though we note that our bootstrapping technique differs from theirs and our test set population of 59 does not meet their minimum population requirements of 123--245.

\begin{table*}[!t]
    \centering
    \caption{Biometric authentication results using bootstrapped similarity score distributions. Results are reported as $\text{mean}_{\pm \text{SD}}$ across 1000~bootstrapped distributions. Each bootstrapped distribution contains $\text{P}=20000$~positives and $\text{N}=20000$~negatives.}
    \label{tab:results-bootstrap}
    \begin{tabular}{@{} c ccrrrr@{}}
        \toprule
        {\multirow{2}[2]{*}{Sampling rate (Hz)}} & {\multirow{2}[2]{*}{Duration (s)}} &  \multirow{2}[2]{*}{EER (\%)} & \multicolumn{4}{c}{FRR @ FAR (\%)} \\ \cmidrule(lr){4-7}
        {} & {} & {} & \multicolumn{1}{c}{$10^{-1}$} & \multicolumn{1}{c}{$10^{-2}$} & \multicolumn{1}{c}{$10^{-3}$} & \multicolumn{1}{c}{$10^{-4}$} \\
        \midrule
        \multirow{5}{*}{\tablenum[table-format=4.2]{1000}} & $5\times\hphantom{1}1$ & $3.67_{\pm 0.12}$ & $3.45_{\pm 0.13}$ & $10.34_{\pm 0.22}$ & $14.27_{\pm 1.04}$ & $30.19_{\pm 2.87}$ \\
        {} & $5\times\hphantom{1}2$ & $2.22_{\pm 0.11}$ & $0.00_{\pm 0.00}$ & $5.16_{\pm 0.19}$ & $13.60_{\pm 1.78}$ & $17.13_{\pm 0.50}$ \\
        {} & $5\times\hphantom{1}4$ & $0.38_{\pm 0.05}$ & $0.00_{\pm 0.00}$ & $0.00_{\pm 0.00}$ & $8.33_{\pm 0.73}$ & $8.48_{\pm 0.20}$ \\
        {} & $5\times\hphantom{1}6$ & $0.56_{\pm 0.05}$ & $0.00_{\pm 0.00}$ & $0.00_{\pm 0.00}$ & $5.08_{\pm 0.16}$ & $5.08_{\pm 0.16}$ \\
        {} & $5\times12$ & $0.59_{\pm 0.05}$ & $0.00_{\pm 0.00}$ & $0.00_{\pm 0.00}$ & $5.09_{\pm 0.16}$ & $5.09_{\pm 0.16}$ \\
        \midrule
        \tablenum[table-format=4.2]{500} & \multirow{5}{*}{$5\times12$} & $0.32_{\pm 0.04}$ & $0.00_{\pm 0.00}$ & $0.00_{\pm 0.00}$ & $3.82_{\pm 0.31}$ & $7.46_{\pm 0.43}$ \\
        \tablenum[table-format=4.2]{250} & {} & $0.81_{\pm 0.06}$ & $0.00_{\pm 0.00}$ & $0.00_{\pm 0.00}$ & $3.78_{\pm 0.13}$ & $5.58_{\pm 0.44}$ \\
        \tablenum[table-format=4.2]{125} & {} & $3.49_{\pm 0.12}$ & $0.00_{\pm 0.00}$ & $7.02_{\pm 0.18}$ & $10.47_{\pm 0.36}$ & $10.52_{\pm 0.21}$ \\
        \tablenum[table-format=4.2]{50} & {} & $3.38_{\pm 0.13}$ & $0.00_{\pm 0.00}$ & $10.40_{\pm 0.63}$ & $16.06_{\pm 1.47}$ & $20.21_{\pm 0.52}$ \\
        \tablenum[table-format=4.2]{31.25} & {} & $5.09_{\pm 0.16}$ & $5.09_{\pm 0.16}$ & $14.47_{\pm 1.12}$ & $29.36_{\pm 2.12}$ & $51.28_{\pm 4.26}$ \\
        \bottomrule
    \end{tabular}
\end{table*}

\begin{figure}[!t]
    \centering
    \includegraphics[width=\linewidth]{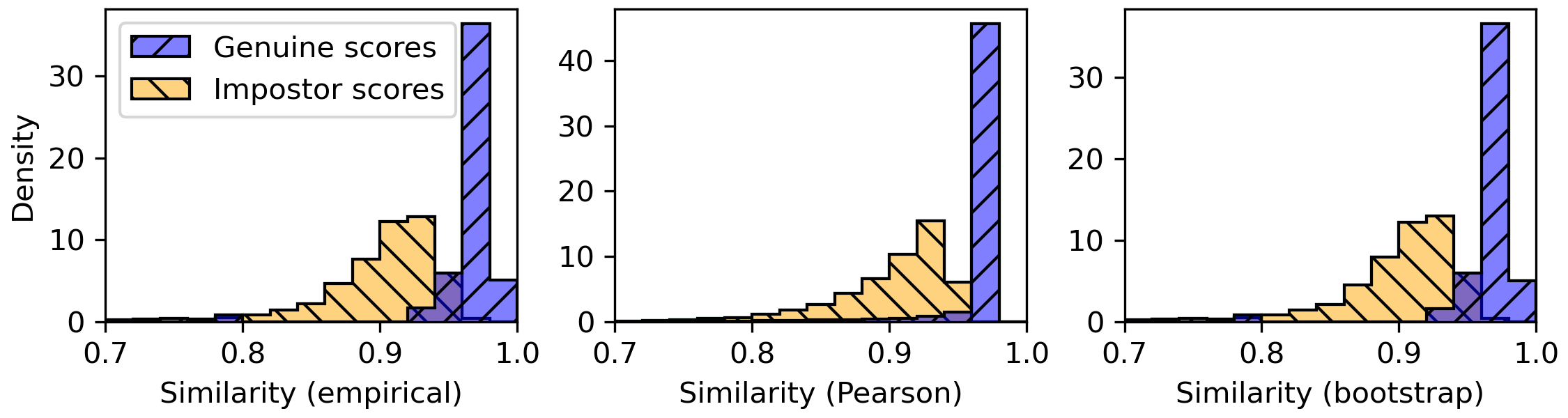}
    \caption{Comparison of genuine vs impostor similarity score distributions for (left) the empirical distributions, (center) resampled distributions using the Pearson-based method from~\cite{lohr2021eye}, and (right) resampled distributions using bootstrapping. We note that bootstrapping more closely preserves the shape of the empirical distributions, particularly the overlapping region between the genuine and impostor distributions. In contrast, the Pearson-based method produces significantly less overlap between the genuine and impostor distributions, leading to a significant reduction in FRR @ FAR $10^{-4}$ compared to the bootstrapped distributions. Plotted scores are from evaluating our model on 31.25~Hz R1 TEX with $5\times12$ inputs. In this example, the Pearson-based method results in 6.26\% FRR @ FAR $10^{-4}$ which is significantly different from the bootstrapped result of 51.28\%.}
    \label{fig:pears-fail}
\end{figure}

We note that \ac{eky}~\cite{lohr2021eye} employs a different method to estimate FRR @ FAR $10^{-4}$ involving the Pearson family of distributions.
We propose the use of bootstrapping because it is simpler, makes fewer assumptions about the empirical distribution, and is more commonly used as a resampling tool.
Bootstrapping largely preserves the shape of the empirical distribution, whereas the Pearson-based approach by~\cite{lohr2021eye} produces a new distribution that may not preserve characteristics of the region of interest where the genuine and impostor distributions overlap.
A ``failure case'' of the Pearson-based approach is shown in Fig.~\ref{fig:pears-fail}.
We believe the main source of this failure is that the empirical distribution of genuine scores is not unimodal (there appears to be a smaller second mode around 0.8~similarity), so we break the assumptions of the Pearson distribution.

\subsection{Determining accept/reject threshold on validation set}
\noindent As mentioned in~\cite{lohr2021eye}, it is problematic to compute \ac{eer} on the test set, because doing so leaks information from the test set into the decision of which accept/reject threshold to use.
A more principled approach is to use the validation set to determine the accept/reject threshold and then apply that threshold to the test set.

For this analysis, we do exactly that.
For each individual model from the ensemble, we build a ROC curve using similarity scores computed on that model's validation set and determine the threshold that yields the \ac{eer}.
Then, separately for each model, we apply the chosen threshold on the similarity scores from the test set.

Note that for this analysis, unlike the previous analyses, we are no longer treating the 4~models as a single ensemble model, because each model's validation set is present in the train set for the other 3~models.
So, to provide a better understanding of performance without ensembling the individual models, we also measure \ac{eer} for each individual model.

Results for this analysis are presented in Table~\ref{tab:results-val-fit}.
We label the folds F0, F1, F2, and F3 and match each model to the fold that was used as its validation set.

\begin{table}[!t]
    \centering
    \caption{Biometric authentication results on the test set for each individual model from the ensemble when the accept/reject threshold is determined on either the test set or a given model's validation set.}
    \label{tab:results-val-fit}
    \begin{tabular}{@{}c S[table-format=1.4] S[table-format=1.2] S[table-format=1.4] S[table-format=2.2] S[table-format=1.2] @{}}
        \toprule
        \multirow{2}[2]{*}{Fold} & \multicolumn{2}{c}{{Fit on test set}} & \multicolumn{3}{c}{{Fit on validation set}}\\ \cmidrule(lr){2-3} \cmidrule(lr){4-6}
        {} & {threshold} & {EER (\%)} & {threshold} & {FRR (\%)} & {FAR (\%)} \\
        \midrule
        F0 & 0.4231 & 5.71 & 0.5405 & 12.07 & 1.28 \\
        F1 & 0.3970 & 8.62 & 0.5850 & 15.52 & 0.65 \\
        F2 & 0.4505 & 5.17 & 0.5575 & 10.34 & 1.25 \\
        F3 & 0.4277 & 6.90 & 0.5024 & 8.62 & 2.82 \\
        \bottomrule
    \end{tabular}
\end{table}

\subsection{Comparison to previous state of the art}
\label{sec:comparison}
\noindent The statistical approach by Friedman et al.~\cite{Friedman2017} reports 2.01\% EER ($\text{P}=149$, $\text{N}=22052$) on the TEX task from GazeBase when enrolling with the full duration (approx. 60~s) of the first session of R1 and authenticating with the full duration (approx. 60~s) of the second session of R1.
This result can be directly compared to our result of 0.58\% EER ($\text{P}=59$, $\text{N}=3422$) for the $5\times12$ duration in Table~\ref{tab:main-results-noexclude-train-drop50} in the ``Duration'' group.

\ac{eky}~\cite{lohr2021eye} reports 14.88\% EER ($\text{P}=59$, $\text{N}=3422$) on the TEX task from GazeBase when enrolling with the first 5~s of the first session of R1 and authenticating with the first 5~s of the second session of R1.
This result can be directly compared to our primary result in the first row of Table~\ref{tab:main-results-noexclude-train-drop50} where we achieve 3.66\% EER ($\text{P}=58$, $\text{N}=3364$) under the same condition (except we exclude windows with more than 50\% NaNs).

\ac{del}~\cite{makowski2021deepeyedentificationlive} reports 10.0\% EER ($\text{P}=25$, $\text{N}=600$) on the TEX task from GazeBase when enrolling with 24~s of data sampled from R1--2 and authenticating with 5~s of data sampled from R3--4.
This result can be roughly compared to our results in Table~\ref{tab:main-results-noexclude-train-drop50} in the ``Test-retest interval'' group, particularly for R2--4, where we achieve between 7.43\% and 8.71\% EER ($\text{P}=58$, $\text{N}=3364$).
Though, we note that we enroll with only the first 5~s of the first session of R1.

It is also worth briefly comparing against \ac{eky} and \ac{del} at degraded sampling rates.
\ac{del} reports a mean EER of 9\% with 5~s of 125~Hz data in~\cite{prasse2020relationship}.
This is similar to our result of 8.77\% EER with 5~s of 125~Hz data, but it is difficult to directly compare across studies since we use a different dataset.
\ac{eky} reports a mean EER of 18.75\% with 10~s of 125~Hz data which is significantly worse than our results.
We note that estimates of FRR @ FAR are not reported in~\cite{prasse2020relationship}.

For our final analysis, we evaluate our model on the JuDo1000~\cite{Makowski2020} dataset which is recorded with an eye-tracking device similar to the one used in GazeBase and which uses an eye-tracking task similar to the RAN task from GazeBase.
JuDo1000 contains 150~subjects recorded across 4~sessions, each separated by at least 1~week.
Each session contains 12~repetitions each of 9~different trial configurations.
More details about the dataset can be found in~\cite{Makowski2020}.

From each recording, the 12~trials with the largest display area (grid = 0.25) and longest duration (dur = 1000) are selected, providing us with 12~windows of 5~s each.
JuDo1000 is a binocular dataset but our model was trained on monocular data, so we combine the left and right eye gaze positions by averaging them.
Gaze positions in each window are converted from pixels to degrees and then from position to velocity using the same Savitzky-Golay differentiation filter we used for GazeBase.
Using the same trained model that produced the results shown in Table~\ref{tab:main-results-noexclude-train-drop50} and without any fine-tuning, we simply compute embeddings of these windows from JuDo1000.
We enroll the embeddings from the first recording session and authenticate with the embeddings from the second recording session (a test-retest interval of approximately 1~week), excluding windows with more than 50\% NaNs.

Table~\ref{tab:results-judo} shows our results alongside the published results of \ac{del}.
Since there are more than 10000~negative pairs, FRR @ FAR $10^{-4}$ is directly measured without any resampling.
We note that this is not a one-to-one comparison with \ac{del} for several reasons, including: \ac{del} trained on JuDo1000 but we trained on GazeBase; \ac{del} used a total of 12~input channels but we used 2; \ac{del} enrolled using 24~s of data across 3~recording sessions but we enrolled with 5~s from only 1~session; and \ac{del} averages results across all 9~trial configurations in JuDo1000 but we only use 1~trial configuration.
That said, these results show the robustness of our model to completely out-of-sample data.

\begin{table}[!t]
    \centering
    \caption{Biometric authentication results on the JuDo1000~\cite{Makowski2020} dataset.}
    \label{tab:results-judo}
    \begin{tabular}{@{}lc S[table-format=2.2] S[table-format=2.2] S[table-format=2.2] S[table-format=3.0] S[table-format=5.0] @{}}
        \toprule
        \multirow{2}[2]{*}{Model} & \multirow{2}[2]{*}{Duration (s)} & {\multirow{2}[2]{*}{EER (\%)}} & \multicolumn{2}{c}{FRR @ FAR (\%)} & {\multirow{2}[2]{*}{P}} & {\multirow{2}[2]{*}{N}} \\ \cmidrule(lr){4-5}
        {} & {} & {} & {$10^{-2}$} & {$10^{-4}$} & {} & {} \\
        \midrule
        \ac{del}~\cite{makowski2021deepeyedentificationlive} & $1\times\hphantom{1}5$ & 3.97 & 25.67 & {-} & 25 & 600 \\
        {} & $1\times10$ & 3.01 & 22.01 & {-} & 25 & 600 \\
        \midrule
        Ours* & $5\times\hphantom{1}1$ & 12.00 & 44.00 & 92.00 & 150 & 22350 \\
        {} & $5\times\hphantom{1}2$ & 7.33 & 28.67 & 68.67 & 150 & 22350 \\
        {} & $5\times12$ & 2.67 & 5.33 & 24.00 & 150 & 22350 \\
        \bottomrule
        \multicolumn{7}{l}{* Our model was trained on GazeBase and evaluated on JuDo1000.} \\
    \end{tabular}
\end{table}

%% file: sections/sec_06discuss.tex
\section{Discussion}
\label{sec:discussion}
\noindent The primary result that we present is 3.66\% \ac{eer} on a reading task with 5-second-long enrollment and authentication periods and with an approximately 30-minute test-retest interval.
\Ac{eky}~\cite{lohr2021eye} reports a mean \ac{eer} of 14.88\% under similar conditions.
\Ac{del}~\cite{makowski2021deepeyedentificationlive} reports a mean \ac{eer} of 3.97\% on a jumping dot task with an enrollment period lasting 24~seconds over a 3-week period and a 5-second-long authentication period with an approximately 1-week test-retest interval.

Authentication accuracy is generally better for TEX than the other tasks, as is expected given the literature's predominant use of reading data for \ac{emb}.
The \ac{eer} for VD2 is slightly lower than for TEX, but this difference may not be statistically significant and the trend may not continue for longer durations.
Unsurprisingly, authentication accuracy is the worst for FXS; but it is impressive that we manage to achieve below 10\% \ac{eer} given just 5~s of pure fixational data.
We note that the FXS task was not well represented in the training set, because the task has a maximum duration of approx. 15~s (compared to 60--100~s for the other tasks) and all other tasks are likely to elicit several saccadic movements in any given 5-second period.
What is quite surprising, however, is that our model achieves 5.49\% EER on BLG despite that task not being present during training.
BLG presumably elicits very different eye movement responses than the other tasks because it is an interactive game with many objects moving on the screen at once, but our model is still able to create meaningful embeddings of the eye movement signals.
We also draw attention to the fact that the embedding space (Fig.~\ref{fig:densmap-all}) appears to be fairly well-clustered across tasks, suggesting that it may be viable to enroll with one task and authenticate with another.

Our model exhibits high robustness to template aging, even with just 5~s of eye movement data.
When authenticating on R6, which is approx. 1~year after R1 and is not represented in the train or validation sets, we still achieve 6.09\% \ac{eer}.
In fact, \ac{eer} remains consistently between 6--9\% for all test-retest intervals from approx. 1~to 37~months.

It is still an open question as to how much eye movement data is necessary to adequately perform user authentication and whether there is a point beyond which additional data provides no new information.
For our model, \ac{eer} improves as the duration of enrollment and authentication increases from 5~s to 20~s, after which it starts to saturate around 0.4--0.6\%.
Estimates of FRR @ FAR $10^{-4}$ improve with increasing duration up to 30~s before saturating around 5\%.
Our results suggest that there may not be much additional information to be gained beyond 30~s of eye movements during a reading task.
Though, it must be noted that this claim is based on the TEX task from GazeBase wherein each subject read through each passage at different speeds.
Perhaps the reason we do not see much improvement beyond 30~s is that most subjects may have finished reading after 30~s and did not have consistent behavior afterward.

Authentication accuracy remains relatively stable as the sampling rate is degraded from 1000~Hz down to 250~Hz, starts to noticeably worsen at 125~Hz, and then drops significantly starting at 50~Hz.
It is unclear how much of this performance degradation is due to the use of an untuned differentiation filter.
These results may reflect findings in the literature that saccade characteristics (e.g., peak velocity and duration) can be measured accurately at a sampling rate of 250~Hz~\cite{geest2002recording} and begin to become less accurate at lower sampling rates~\cite{andersson2010sampling,leube2017sampling}.
At 125~Hz, which is close to the 120~Hz sampling rate of the Vive Pro Eye~\cite{ViveProEye}, our model is still able to achieve 8.77\% \ac{eer} with just 5~s of data and an estimated 10.52\% FRR @ FAR $10^{-4}$ with 60~s of data.
Depending on the degree of security necessary, these results suggest the present applicability of \ac{emb} at sampling rates present in current VR/AR devices.
Another interesting observation is that we are able to achieve around 5.09\% EER with 60~s of 31.25~Hz data, suggesting that there is still meaningful biometric information that can be extracted at such low sampling rates.
While we acknowledge that simply degrading the sampling rate of high-quality data is not a sufficient proxy for other eye-tracking devices, we note that gaze estimation pipelines could always be improved to produce higher levels of signal quality at a particular sampling rate, whereas it may not always be possible for a device to increase the sampling rate of its eye-tracking sensor(s) due to power constraints (though some efforts are being made to enable eye tracking at high sampling rates with lower power requirements~\cite{stoffregen2022event}).

Most \ac{emb} studies, including the present study, report measures of \ac{eer} directly on the test set, leaking information from the test set into the accept/reject decisions.
When taking a more principled approach and fitting the accept/reject threshold on the validation set instead, we find that the thresholds become more strict, resulting in a lower \ac{far} and a higher \ac{frr}.
As such, these thresholds would be better for settings requiring a higher degree of security but may be more frustrating for users.

%% file: sections/sec_07conclusion.tex
\section{Conclusion}
\label{sec:conclusion}
\noindent We presented a novel, highly parameter-efficient, \ac{densenet}-based architecture for end-to-end \ac{emb} that achieves state-of-the-art biometric authentication performance.
When enrolling and authenticating with just 5~s of eye movements during a reading task---a duration somewhat comparable to the time it takes to enter a 4-digit pin or to calibrate an eye-tracking device---we achieved 3.66\% \ac{eer}.
With 30~s of data, we achieved an estimated 5.08\% FRR @ FAR $10^{-4}$ which approaches a level of authentication performance that would be acceptable for real-world use.
At 125~Hz, which is close to the 120~Hz sampling rate of the Vive Pro Eye~\cite{ViveProEye}, we achieved 8.77\% \ac{eer} with just 5~s of data and an estimated 10.52\% FRR @ FAR $10^{-4}$ with 60~s of data.

Our embedding space visualizations suggest that it may be feasible to enroll with one (or several) tasks and authenticate with a different task.
We are not aware of any study that has attempted this.
It would also be interesting to see how well privacy-preserving models (e.g.,~\cite{david-john2021privacy}) can defend against more powerful \ac{emb} models like the one presented herein.

Since eye tracking is seeing increasing use in VR/AR devices due in part to the power-saving potential of foveated rendering, \ac{emb} may be an ideal biometric modality for such devices.
\Ac{emb} models would need to have low resource requirements when performing (continuous) user authentication on such consumer-grade devices.
Our architecture (excluding the classification layer) has only 123K~learnable parameters which is around 4x~smaller than \ac{eky} (approx. 475K~learnable parameters) and around 1700x~smaller than \ac{del} (approx. 209M~learnable parameters according to~\cite{lohr2021eye}).
Models with lower complexity such as ours may enable more power-efficient implementations that would make them a better fit for deployment on consumer-grade devices.
Following works like~\cite{lohr2020vr}, we encourage future studies to explore \ac{emb} directly on eye-tracking-enabled VR/AR devices.

%% file: 01main.bbl
\begin{thebibliography}{10}
\providecommand{\url}[1]{#1}
\csname url@samestyle\endcsname
\providecommand{\newblock}{\relax}
\providecommand{\bibinfo}[2]{#2}
\providecommand{\BIBentrySTDinterwordspacing}{\spaceskip=0pt\relax}
\providecommand{\BIBentryALTinterwordstretchfactor}{4}
\providecommand{\BIBentryALTinterwordspacing}{\spaceskip=\fontdimen2\font plus
\BIBentryALTinterwordstretchfactor\fontdimen3\font minus
  \fontdimen4\font\relax}
\providecommand{\BIBforeignlanguage}[2]{{%
\expandafter\ifx\csname l@#1\endcsname\relax
\typeout{** WARNING: IEEEtran.bst: No hyphenation pattern has been}%
\typeout{** loaded for the language `#1'. Using the pattern for}%
\typeout{** the default language instead.}%
\else
\language=\csname l@#1\endcsname
\fi
#2}}
\providecommand{\BIBdecl}{\relax}
\BIBdecl

\bibitem{jain2004introduction}
A.~Jain, A.~Ross, and S.~Prabhakar, ``An introduction to biometric
  recognition,'' \emph{IEEE Transactions on Circuits and Systems for Video
  Technology}, vol.~14, no.~1, pp. 4--20, 2004.

\bibitem{Kasprowski2004}
P.~Kasprowski and J.~Ober, ``{Eye movements in biometrics},'' \emph{Lecture
  Notes in Computer Science (including subseries Lecture Notes in Artificial
  Intelligence and Lecture Notes in Bioinformatics)}, vol. 3087, pp. 248--258,
  2004.

\bibitem{Komogortsev2015}
O.~V. {Komogortsev}, A.~{Karpov}, and C.~D. {Holland}, ``Attack of mechanical
  replicas: {Liveness} detection with eye movements,'' \emph{IEEE Transactions
  on Information Forensics and Security}, vol.~10, no.~4, pp. 716--725, 2015.

\bibitem{makowski2021deepeyedentificationlive}
S.~Makowski, P.~Prasse, D.~R. Reich, D.~Krakowczyk, L.~A. Jäger, and
  T.~Scheffer, ``Deepeyedentificationlive: Oculomotoric biometric
  identification and presentation-attack detection using deep neural
  networks,'' \emph{IEEE Transactions on Biometrics, Behavior, and Identity
  Science}, vol.~3, no.~4, pp. 506--518, 2021.

\bibitem{Eberz2015}
S.~Eberz, K.~Rasmussen, V.~Lenders, and I.~Martinovic, ``Preventing lunchtime
  attacks: {Fighting} insider threats with eye movement biometrics,'' 2015.

\bibitem{Eberz2019}
\BIBentryALTinterwordspacing
S.~Eberz, G.~Lovisotto, K.~B. Rasmussen, V.~Lenders, and I.~Martinovic, ``28
  {Blinks} {Later}: {Tackling} {Practical} {Challenges} of {Eye} {Movement}
  {Biometrics},'' in \emph{Proceedings of the 2019 {ACM} {SIGSAC} {Conference}
  on {Computer} and {Communications} {Security}}, ser. {CCS} '19.\hskip 1em
  plus 0.5em minus 0.4em\relax London, United Kingdom: Association for
  Computing Machinery, Nov. 2019, pp. 1187--1199. [Online]. Available:
  \url{https://doi.org/10.1145/3319535.3354233}
\BIBentrySTDinterwordspacing

\bibitem{Kasprowski2018}
\BIBentryALTinterwordspacing
P.~Kasprowski and K.~Harezlak, ``\BIBforeignlanguage{en}{Fusion of eye movement
  and mouse dynamics for reliable behavioral biometrics},''
  \emph{\BIBforeignlanguage{en}{Pattern Analysis and Applications}}, vol.~21,
  no.~1, pp. 91--103, Feb. 2018. [Online]. Available:
  \url{https://doi.org/10.1007/s10044-016-0568-5}
\BIBentrySTDinterwordspacing

\bibitem{komogortsev2012multimodal}
O.~V. Komogortsev, A.~Karpov, C.~D. Holland, and H.~P. Proença, ``Multimodal
  ocular biometrics approach: A feasibility study,'' in \emph{2012 IEEE Fifth
  International Conference on Biometrics: Theory, Applications and Systems
  (BTAS)}, 2012, pp. 209--216.

\bibitem{liebling2014privacy}
D.~J. Liebling and S.~Preibusch, ``Privacy considerations for a pervasive eye
  tracking world.''\hskip 1em plus 0.5em minus 0.4em\relax Association for
  Computing Machinery, Inc, 2014, pp. 1169--1177.

\bibitem{liu2019differential}
\BIBentryALTinterwordspacing
A.~Liu, L.~Xia, A.~Duchowski, R.~Bailey, K.~Holmqvist, and E.~Jain,
  ``Differential privacy for eye-tracking data,'' \emph{Proceedings of the 11th
  ACM Symposium on Eye Tracking Research \& Applications}, p.~10, 2019.
  [Online]. Available: \url{https://doi.org/10.1145/3314111.3319823}
\BIBentrySTDinterwordspacing

\bibitem{steil2019privacy}
\BIBentryALTinterwordspacing
J.~Steil, I.~Hagestedt, M.~X. Huang, and A.~Bulling, ``Privacy-aware eye
  tracking using differential privacy.''\hskip 1em plus 0.5em minus 0.4em\relax
  ACM, 6 2019, pp. 1--9. [Online]. Available:
  \url{https://dl.acm.org/doi/10.1145/3314111.3319915}
\BIBentrySTDinterwordspacing

\bibitem{david-john2021privacy}
\BIBentryALTinterwordspacing
B.~David-John, D.~Hosfelt, K.~Butler, and E.~Jain, ``A privacy-preserving
  approach to streaming eye-tracking data,'' \emph{IEEE Transactions on
  Visualization and Computer Graphics}, 2021. [Online]. Available:
  \url{https://www.nytimes.com/interactive/2019/12/19/opinion/}
\BIBentrySTDinterwordspacing

\bibitem{li2021kaleido}
J.~Li, A.~R. Chowdhury, K.~Fawaz, and Y.~Kim, ``Kaleido: Real-time privacy
  control for eye-tracking systems,'' 2021.

\bibitem{ViveProEye}
``{Vive Pro Eye},''
  \url{https://www.vive.com/us/product/vive-pro-eye/overview/}, accessed:
  2021-04-07.

\bibitem{ML1}
``{Magic Leap} 1,'' \url{https://www.magicleap.com/en-us/magic-leap-1},
  accessed: 2021-04-07.

\bibitem{HL2}
``{HoloLens} 2,'' \url{https://www.microsoft.com/en-us/hololens}, accessed:
  2022-03-27.

\bibitem{guenter2012foveated}
\BIBentryALTinterwordspacing
B.~Guenter, M.~Finch, S.~Drucker, D.~Tan, and J.~Snyder, ``Foveated 3d
  graphics,'' \emph{ACM Trans. Graph.}, vol.~31, no.~6, nov 2012. [Online].
  Available: \url{https://doi.org/10.1145/2366145.2366183}
\BIBentrySTDinterwordspacing

\bibitem{stoffregen2022event}
T.~Stoffregen, H.~Daraei, C.~Robinson, and A.~Fix, ``Event-based kilohertz eye
  tracking using coded differential lighting,'' in \emph{2022 IEEE/CVF Winter
  Conference on Applications of Computer Vision (WACV)}, 2022, pp. 3937--3945.

\bibitem{lohr2020vr}
\BIBentryALTinterwordspacing
D.~J. Lohr, S.~Aziz, and O.~Komogortsev, ``Eye movement biometrics using a new
  dataset collected in virtual reality,'' in \emph{ACM Symposium on Eye
  Tracking Research and Applications}, ser. ETRA ’20 Adjunct.\hskip 1em plus
  0.5em minus 0.4em\relax New York, NY, USA: Association for Computing
  Machinery, 2020. [Online]. Available:
  \url{https://doi.org/10.1145/3379157.3391420}
\BIBentrySTDinterwordspacing

\bibitem{Jia2018}
S.~Jia, D.~H. Koh, A.~Seccia, P.~Antonenko, R.~Lamb, A.~Keil, M.~Schneps, and
  M.~Pomplun, ``{Biometric recognition through eye movements using a recurrent
  neural network},'' in \emph{Proceedings - 9th IEEE International Conference
  on Big Knowledge, ICBK 2018}.\hskip 1em plus 0.5em minus 0.4em\relax
  Institute of Electrical and Electronics Engineers Inc., dec 2018, pp. 57--64.

\bibitem{Abdelwahab2019}
\BIBentryALTinterwordspacing
A.~Abdelwahab and N.~Landwehr, ``Deep {Distributional} {Sequence} {Embeddings}
  {Based} on a {Wasserstein} {Loss},'' \emph{arXiv:1912.01933 [cs, stat]}, Dec.
  2019, arXiv: 1912.01933. [Online]. Available:
  \url{http://arxiv.org/abs/1912.01933}
\BIBentrySTDinterwordspacing

\bibitem{Jager2020}
L.~A. J{\"a}ger, S.~Makowski, P.~Prasse, S.~Liehr, M.~Seidler, and T.~Scheffer,
  ``Deep eyedentification: Biometric identification using micro-movements of
  the eye,'' in \emph{Machine Learning and Knowledge Discovery in Databases},
  U.~Brefeld, E.~Fromont, A.~Hotho, A.~Knobbe, M.~Maathuis, and C.~Robardet,
  Eds.\hskip 1em plus 0.5em minus 0.4em\relax Cham: Springer International
  Publishing, 2020, pp. 299--314.

\bibitem{lohr2021eye}
D.~Lohr, H.~Griffith, and O.~V. Komogortsev, ``Eye know you: Metric learning
  for end-to-end biometric authentication using eye movements from a
  longitudinal dataset,'' 2021.

\bibitem{he2015deep}
K.~He, X.~Zhang, S.~Ren, and J.~Sun, ``Deep residual learning for image
  recognition,'' 2015.

\bibitem{huang2018densely}
G.~Huang, Z.~Liu, L.~van~der Maaten, and K.~Q. Weinberger, ``Densely connected
  convolutional networks,'' 2018.

\bibitem{griffith2021gazebase}
\BIBentryALTinterwordspacing
H.~Griffith, D.~Lohr, E.~Abdulin, and O.~Komogortsev, ``Gazebase, a
  large-scale, multi-stimulus, longitudinal eye movement dataset,''
  \emph{Scientific Data}, vol.~8, no.~1, p. 184, Jul 2021. [Online]. Available:
  \url{https://doi.org/10.1038/s41597-021-00959-y}
\BIBentrySTDinterwordspacing

\bibitem{Friedman2020}
L.~Friedman, H.~S. Stern, V.~Prokopenko, S.~Djanian, H.~K. Griffith, and O.~V.
  Komogortsev, ``Biometric performance as a function of gallery size,'' 2020.

\bibitem{unar2014review}
\BIBentryALTinterwordspacing
J.~Unar, W.~C. Seng, and A.~Abbasi, ``A review of biometric technology along
  with trends and prospects,'' \emph{Pattern Recognition}, vol.~47, no.~8, pp.
  2673--2688, 2014. [Online]. Available:
  \url{https://www.sciencedirect.com/science/article/pii/S003132031400034X}
\BIBentrySTDinterwordspacing

\bibitem{alex2012imagenet}
\BIBentryALTinterwordspacing
A.~Krizhevsky, I.~Sutskever, and G.~E. Hinton, ``Imagenet classification with
  deep convolutional neural networks,'' in \emph{Advances in Neural Information
  Processing Systems}, F.~Pereira, C.~J.~C. Burges, L.~Bottou, and K.~Q.
  Weinberger, Eds., vol.~25.\hskip 1em plus 0.5em minus 0.4em\relax Curran
  Associates, Inc., 2012. [Online]. Available:
  \url{https://proceedings.neurips.cc/paper/2012/file/c399862d3b9d6b76c8436e924a68c45b-Paper.pdf}
\BIBentrySTDinterwordspacing

\bibitem{simonyan2015deep}
K.~Simonyan and A.~Zisserman, ``Very deep convolutional networks for
  large-scale image recognition,'' 2015.

\bibitem{elmadjian2021tcn}
C.~Elmadjian, C.~Gonzales, and C.~H. Morimoto, ``Eye movement classification
  with temporal convolutional networks,'' in \emph{Pattern Recognition. ICPR
  International Workshops and Challenges}, A.~Del~Bimbo, R.~Cucchiara,
  S.~Sclaroff, G.~M. Farinella, T.~Mei, M.~Bertini, H.~J. Escalante, and
  R.~Vezzani, Eds.\hskip 1em plus 0.5em minus 0.4em\relax Cham: Springer
  International Publishing, 2021, pp. 390--404.

\bibitem{oord2016wavenet}
\BIBentryALTinterwordspacing
A.~van~den Oord, S.~Dieleman, H.~Zen, K.~Simonyan, O.~Vinyals, A.~Graves,
  N.~Kalchbrenner, A.~Senior, and K.~Kavukcuoglu, ``Wavenet: A generative model
  for raw audio,'' in \emph{Arxiv}, 2016. [Online]. Available:
  \url{https://arxiv.org/abs/1609.03499}
\BIBentrySTDinterwordspacing

\bibitem{hochreiter1997lstm}
\BIBentryALTinterwordspacing
S.~Hochreiter and J.~Schmidhuber, ``{Long Short-Term Memory},'' \emph{Neural
  Computation}, vol.~9, no.~8, pp. 1735--1780, 11 1997. [Online]. Available:
  \url{https://doi.org/10.1162/neco.1997.9.8.1735}
\BIBentrySTDinterwordspacing

\bibitem{cho2014learning}
K.~Cho, B.~van Merrienboer, C.~Gulcehre, D.~Bahdanau, F.~Bougares, H.~Schwenk,
  and Y.~Bengio, ``Learning phrase representations using rnn encoder-decoder
  for statistical machine translation,'' 2014.

\bibitem{lea2016temporal}
C.~Lea, R.~Vidal, A.~Reiter, and G.~D. Hager, ``Temporal convolutional
  networks: A unified approach to action segmentation,'' 2016.

\bibitem{li2018visualizing}
H.~Li, Z.~Xu, G.~Taylor, C.~Studer, and T.~Goldstein, ``Visualizing the loss
  landscape of neural nets,'' 2018.

\bibitem{xie2017aggregated}
S.~Xie, R.~Girshick, P.~Dollár, Z.~Tu, and K.~He, ``Aggregated residual
  transformations for deep neural networks,'' 2017.

\bibitem{zhang2020resnet}
C.~Zhang, P.~Benz, D.~M. Argaw, S.~Lee, J.~Kim, F.~Rameau, J.-C. Bazin, and
  I.~S. Kweon, ``Resnet or densenet? introducing dense shortcuts to resnet,''
  2020.

\bibitem{tan2020efficientnet}
M.~Tan and Q.~V. Le, ``Efficientnet: Rethinking model scaling for convolutional
  neural networks,'' 2020.

\bibitem{tan2021efficientnetv2}
------, ``Efficientnetv2: Smaller models and faster training,'' 2021.

\bibitem{vaswani2017attention}
\BIBentryALTinterwordspacing
A.~Vaswani, N.~Shazeer, N.~Parmar, J.~Uszkoreit, L.~Jones, A.~N. Gomez, L.~u.
  Kaiser, and I.~Polosukhin, ``Attention is all you need,'' in \emph{Advances
  in Neural Information Processing Systems}, I.~Guyon, U.~V. Luxburg,
  S.~Bengio, H.~Wallach, R.~Fergus, S.~Vishwanathan, and R.~Garnett, Eds.,
  vol.~30.\hskip 1em plus 0.5em minus 0.4em\relax Curran Associates, Inc.,
  2017. [Online]. Available:
  \url{https://proceedings.neurips.cc/paper/2017/file/3f5ee243547dee91fbd053c1c4a845aa-Paper.pdf}
\BIBentrySTDinterwordspacing

\bibitem{dosovitskiy2021vit}
\BIBentryALTinterwordspacing
A.~Dosovitskiy, L.~Beyer, A.~Kolesnikov, D.~Weissenborn, X.~Zhai,
  T.~Unterthiner, M.~Dehghani, M.~Minderer, G.~Heigold, S.~Gelly, J.~Uszkoreit,
  and N.~Houlsby, ``An image is worth 16x16 words: Transformers for image
  recognition at scale,'' in \emph{International Conference on Learning
  Representations}, 2021. [Online]. Available:
  \url{https://openreview.net/forum?id=YicbFdNTTy}
\BIBentrySTDinterwordspacing

\bibitem{holland2011biometric}
C.~Holland and O.~V. Komogortsev, ``Biometric identification via eye movement
  scanpaths in reading,'' in \emph{2011 International Joint Conference on
  Biometrics (IJCB)}, 2011, pp. 1--8.

\bibitem{george2016score}
\BIBentryALTinterwordspacing
A.~George and A.~Routray, ``A score level fusion method for eye movement
  biometrics,'' \emph{Pattern Recogn. Lett.}, vol.~82, no.~P2, p. 207–215,
  oct 2016. [Online]. Available:
  \url{https://doi.org/10.1016/j.patrec.2015.11.020}
\BIBentrySTDinterwordspacing

\bibitem{Friedman2017}
\BIBentryALTinterwordspacing
L.~Friedman, M.~S. Nixon, and O.~V. Komogortsev, ``Method to assess the
  temporal persistence of potential biometric features: Application to
  oculomotor, gait, face and brain structure databases,'' \emph{PLOS ONE},
  vol.~12, no.~6, pp. 1--42, 06 2017. [Online]. Available:
  \url{https://doi.org/10.1371/journal.pone.0178501}
\BIBentrySTDinterwordspacing

\bibitem{lohr2020metric}
D.~Lohr, H.~Griffith, S.~Aziz, and O.~Komogortsev, ``A metric learning approach
  to eye movement biometrics,'' in \emph{2020 IEEE International Joint
  Conference on Biometrics (IJCB)}, 2020, pp. 1--7.

\bibitem{porta2022gaze}
M.~Porta, P.~Dondi, N.~Zangrandi, and L.~Lombardi, ``Gaze-based biometrics from
  free observation of moving elements,'' \emph{IEEE Transactions on Biometrics,
  Behavior, and Identity Science}, vol.~4, no.~1, pp. 85--96, 2022.

\bibitem{ioffe2015batchnorm}
\BIBentryALTinterwordspacing
S.~Ioffe and C.~Szegedy, ``Batch normalization: Accelerating deep network
  training by reducing internal covariate shift,'' in \emph{Proceedings of the
  32nd International Conference on Machine Learning}, ser. Proceedings of
  Machine Learning Research, F.~Bach and D.~Blei, Eds., vol.~37.\hskip 1em plus
  0.5em minus 0.4em\relax Lille, France: PMLR, 07--09 Jul 2015, pp. 448--456.
  [Online]. Available: \url{https://proceedings.mlr.press/v37/ioffe15.html}
\BIBentrySTDinterwordspacing

\bibitem{nair2010relu}
V.~Nair and G.~E. Hinton, ``Rectified linear units improve restricted boltzmann
  machines,'' in \emph{Proceedings of the 27th International Conference on
  International Conference on Machine Learning}, ser. ICML'10.\hskip 1em plus
  0.5em minus 0.4em\relax Madison, WI, USA: Omnipress, 2010, p. 807–814.

\bibitem{he2015delving}
\BIBentryALTinterwordspacing
K.~He, X.~Zhang, S.~Ren, and J.~Sun, ``Delving deep into rectifiers: Surpassing
  human-level performance on imagenet classification,'' 2015. [Online].
  Available: \url{https://arxiv.org/abs/1502.01852}
\BIBentrySTDinterwordspacing

\bibitem{wan2019reconciling}
K.~Wan, S.~Yang, B.~Feng, Y.~Ding, and L.~Xie, ``Reconciling feature-reuse and
  overfitting in densenet with specialized dropout,'' in \emph{2019 IEEE 31st
  International Conference on Tools with Artificial Intelligence (ICTAI)},
  2019, pp. 760--767.

\bibitem{he2016identity}
K.~He, X.~Zhang, S.~Ren, and J.~Sun, ``Identity mappings in deep residual
  networks,'' 2016.

\bibitem{paszke2019pytorch}
\BIBentryALTinterwordspacing
A.~Paszke, S.~Gross, F.~Massa, A.~Lerer, J.~Bradbury, G.~Chanan, T.~Killeen,
  Z.~Lin, N.~Gimelshein, L.~Antiga, A.~Desmaison, A.~Kopf, E.~Yang, Z.~DeVito,
  M.~Raison, A.~Tejani, S.~Chilamkurthy, B.~Steiner, L.~Fang, J.~Bai, and
  S.~Chintala, ``Pytorch: An imperative style, high-performance deep learning
  library,'' in \emph{Advances in Neural Information Processing Systems 32},
  H.~Wallach, H.~Larochelle, A.~Beygelzimer, F.~d\textquotesingle
  Alch\'{e}-Buc, E.~Fox, and R.~Garnett, Eds.\hskip 1em plus 0.5em minus
  0.4em\relax Curran Associates, Inc., 2019, pp. 8024--8035. [Online].
  Available:
  \url{http://papers.neurips.cc/paper/9015-pytorch-an-imperative-style-high-performance-deep-learning-library.pdf}
\BIBentrySTDinterwordspacing

\bibitem{Musgrave2020a}
K.~Musgrave, S.~Belongie, and S.-N. Lim, ``Pytorch metric learning,'' 2020.

\bibitem{falcon2019pl}
\BIBentryALTinterwordspacing
W.~Falcon and {The PyTorch Lightning team}, ``{PyTorch Lightning},'' 3 2019.
  [Online]. Available:
  \url{https://github.com/PyTorchLightning/pytorch-lightning}
\BIBentrySTDinterwordspacing

\bibitem{biewald2020wandb}
\BIBentryALTinterwordspacing
L.~Biewald, ``Experiment tracking with weights and biases,'' 2020, software
  available from wandb.com. [Online]. Available: \url{https://www.wandb.com/}
\BIBentrySTDinterwordspacing

\bibitem{mcinnes2018umap-software}
L.~McInnes, J.~Healy, N.~Saul, and L.~Grossberger, ``Umap: Uniform manifold
  approximation and projection,'' \emph{The Journal of Open Source Software},
  vol.~3, no.~29, p. 861, 2018.

\bibitem{savitzky1964sgolay}
\BIBentryALTinterwordspacing
A.~Savitzky and M.~J.~E. Golay, ``Smoothing and differentiation of data by
  simplified least squares procedures.'' \emph{Analytical Chemistry}, vol.~36,
  no.~8, pp. 1627--1639, 1964. [Online]. Available:
  \url{https://doi.org/10.1021/ac60214a047}
\BIBentrySTDinterwordspacing

\bibitem{Wang2019}
X.~{Wang}, X.~{Han}, W.~{Huang}, D.~{Dong}, and M.~R. {Scott},
  ``Multi-similarity loss with general pair weighting for deep metric
  learning,'' in \emph{2019 IEEE/CVF Conference on Computer Vision and Pattern
  Recognition (CVPR)}, 2019, pp. 5017--5025.

\bibitem{kingma2014adam}
D.~P. Kingma and J.~Ba, ``Adam: A method for stochastic optimization,''
  \emph{arXiv preprint arXiv:1412.6980}, 2014.

\bibitem{smith2018superconvergence}
L.~N. Smith and N.~Topin, ``Super-convergence: Very fast training of neural
  networks using large learning rates,'' 2018.

\bibitem{Musgrave2020}
K.~Musgrave, S.~Belongie, and S.-N. Lim, ``A metric learning reality check,''
  in \emph{Computer Vision -- ECCV 2020}, A.~Vedaldi, H.~Bischof, T.~Brox, and
  J.-M. Frahm, Eds.\hskip 1em plus 0.5em minus 0.4em\relax Cham: Springer
  International Publishing, 2020, pp. 681--699.

\bibitem{narayan2020densmap}
\BIBentryALTinterwordspacing
A.~Narayan, B.~Berger, and H.~Cho, ``Density-preserving data visualization
  unveils dynamic patterns of single-cell transcriptomic variability,''
  \emph{bioRxiv}, 2020. [Online]. Available:
  \url{https://www.biorxiv.org/content/early/2020/05/14/2020.05.12.077776}
\BIBentrySTDinterwordspacing

\bibitem{SciPy2020}
P.~Virtanen, R.~Gommers, T.~E. Oliphant, M.~Haberland, T.~Reddy, D.~Cournapeau,
  E.~Burovski, P.~Peterson, W.~Weckesser, J.~Bright, S.~J. {van der Walt},
  M.~Brett, J.~Wilson, K.~J. Millman, N.~Mayorov, A.~R.~J. Nelson, E.~Jones,
  R.~Kern, E.~Larson, C.~J. Carey, {\.I}.~Polat, Y.~Feng, E.~W. Moore,
  J.~{VanderPlas}, D.~Laxalde, J.~Perktold, R.~Cimrman, I.~Henriksen, E.~A.
  Quintero, C.~R. Harris, A.~M. Archibald, A.~H. Ribeiro, F.~Pedregosa, P.~{van
  Mulbregt}, and {SciPy 1.0 Contributors}, ``{{SciPy} 1.0: Fundamental
  Algorithms for Scientific Computing in Python},'' \emph{Nature Methods},
  vol.~17, pp. 261--272, 2020.

\bibitem{FIDO2020}
S.~Schuckers, G.~Cannon, and N.~Tekampe, ``{FIDO} biometrics requirements,''
  \url{https://fidoalliance.org/specs/biometric/requirements/}, accessed:
  2021-04-04.

\bibitem{prasse2020relationship}
P.~Prasse, L.~A. J{\"a}ger, S.~Makowski, M.~Feuerpfeil, and T.~Scheffer, ``On
  the relationship between eye tracking resolution and performance of
  oculomotoric biometric identification,'' vol. 176.\hskip 1em plus 0.5em minus
  0.4em\relax Elsevier, 1 2020, pp. 2088--2097.

\bibitem{Makowski2020}
S.~{Makowski}, L.~A. {Jäger}, P.~{Prasse}, and T.~{Scheffer}, ``Biometric
  identification and presentation-attack detection using micro- and
  macro-movements of the eyes,'' in \emph{2020 IEEE International Joint
  Conference on Biometrics (IJCB)}, 2020, pp. 1--10.

\bibitem{geest2002recording}
\BIBentryALTinterwordspacing
J.~{van der Geest} and M.~Frens, ``Recording eye movements with
  video-oculography and scleral search coils: a direct comparison of two
  methods,'' \emph{Journal of Neuroscience Methods}, vol. 114, no.~2, pp.
  185--195, 2002. [Online]. Available:
  \url{https://www.sciencedirect.com/science/article/pii/S0165027001005271}
\BIBentrySTDinterwordspacing

\bibitem{andersson2010sampling}
\BIBentryALTinterwordspacing
R.~Andersson, M.~Nystr{\"o}m, and K.~Holmqvist, ``Sampling frequency and
  eye-tracking measures: how speed affects durations, latencies, and more,''
  \emph{Journal of Eye Movement Research}, vol.~3, no.~3, Sep. 2010. [Online].
  Available: \url{https://bop.unibe.ch/JEMR/article/view/2300}
\BIBentrySTDinterwordspacing

\bibitem{leube2017sampling}
\BIBentryALTinterwordspacing
A.~Leube and K.~Rifai, ``\BIBforeignlanguage{eng}{Sampling rate influences
  saccade detection in mobile eye tracking of a reading task},''
  \emph{\BIBforeignlanguage{eng}{Journal of eye movement research}}, vol.~10,
  no.~3, p. 10.16910/jemr.10.3.3, Jun 2017, 33828659[pmid]. [Online].
  Available: \url{https://pubmed.ncbi.nlm.nih.gov/33828659}
\BIBentrySTDinterwordspacing

\end{thebibliography}
